\documentclass[10pt,twocolumn,letterpaper]{article}

\usepackage{cvpr}             
\usepackage[accsupp]{axessibility}
\usepackage{graphicx}
\graphicspath{{images/}}
\usepackage{amsmath}
\usepackage{amssymb}
\usepackage{booktabs}
\usepackage{multirow}
\usepackage{amsfonts}
\usepackage{float}
\usepackage{placeins} 
\usepackage{bm}
\usepackage{cite}
\usepackage{siunitx}
\usepackage{amsfonts}
\usepackage[switch]{lineno}
\usepackage[dvipsnames]{xcolor}
\usepackage{enumitem}
\usepackage[pagebackref,breaklinks,colorlinks]{hyperref}
\usepackage[capitalize]{cleveref}
\crefname{section}{Sec.}{Secs.}
\Crefname{section}{Section}{Sections}
\Crefname{table}{Table}{Tables}
\crefname{table}{Tab.}{Tabs.}

\def\cvprPaperID{731} 
\def\confName{CVPR}
\def\confYear{2023}

\begin{document}

\title{Semi-DETR: Semi-Supervised Object Detection with Detection Transformers}

\author{Jiacheng Zhang$^{1,2}$\thanks{Equally-contributed authors. Work done during an internship at Baidu.}\quad
Xiangru Lin$^{2*}$\quad
Wei Zhang$^{2}$\quad
Kuo Wang$^{1}$\quad
Xiao Tan$^{2}$\quad \\
Junyu Han$^{2}$\quad
Errui Ding$^{2}$\quad
Jingdong Wang$^{2}$\quad
Guanbin Li$^{1,3}$\thanks{Corresponding author.}\quad\\
$^{1}$School of Computer Science and Engineering, Sun Yat-sen University, Guangzhou, China\\
$^{2}$Department of Computer Vision Technology (VIS), Baidu Inc., China\\
$^{3}$Research Institute, Sun Yat-sen University, Shenzhen, China \\
{\tt\small \{zhangjch58, wangk229\}@mail2.sysu.edu.cn, liguanbin@mail.sysu.edu.cn } \\
{\tt\small \{linxiangru,zhangwei99,tanxiao01,hanjunyu,dingerrui,wangjingdong\}@baidu.com}
}

\maketitle
\begin{abstract}
We analyze the DETR-based framework on semi-supervised object detection (SSOD) and observe that (1) the one-to-one assignment strategy generates incorrect matching when the pseudo ground-truth bounding box is inaccurate, leading to training inefficiency; (2) DETR-based detectors lack deterministic correspondence between the input query and its prediction output, which hinders the applicability of the consistency-based regularization widely used in current SSOD methods. We present Semi-DETR, the first transformer-based end-to-end semi-supervised object detector, to tackle these problems. Specifically, we propose a Stage-wise Hybrid Matching strategy that combines the one-to-many assignment and one-to-one assignment strategies to improve the training efficiency of the first stage and thus provide high-quality pseudo labels for the training of the second stage. Besides, we introduce a Cross-view Query Consistency method to learn the semantic feature invariance of object queries from different views while avoiding the need to find deterministic query correspondence. Furthermore, we propose a Cost-based Pseudo Label Mining module to dynamically mine more pseudo boxes based on the matching cost of pseudo ground truth bounding boxes for consistency training. Extensive experiments on all SSOD settings of both COCO and Pascal VOC benchmark datasets show that our Semi-DETR method outperforms all state-of-the-art methods by clear margins. The PaddlePaddle version code\footnote{The Pytorch version code is at \url{https://github.com/JCZ404/Semi-DETR}} is at \url{https://github.com/PaddlePaddle/PaddleDetection/tree/develop/configs/semi_det/semi_detr}.

\end{abstract}

\section{Introduction}
\label{sec:intro}

\begin{figure}[ht]
    \centering
    \includegraphics[width=0.9\linewidth]{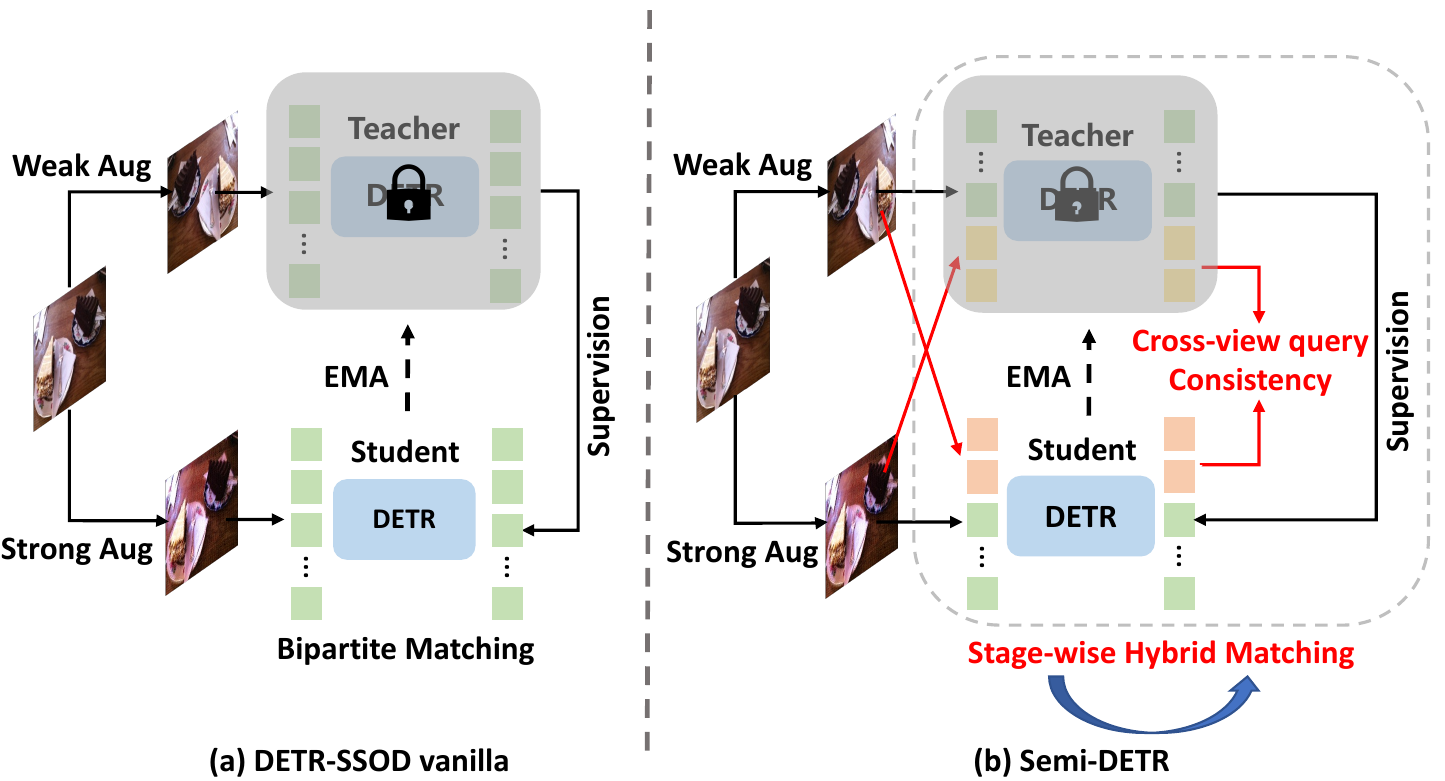}
    \caption{Comparisons between the vanilla DETR-SSOD framework based on the Teacher-Student architecture and our proposed Semi-DETR framework. Semi-DETR consists of Stage-wise Hybrid Matching, Cross-view Query Consistency powered by a cost-based pseudo label mining strategy.}
    \label{fig:semi-detr-intro}
    \vspace{-3mm}
\end{figure}

Semi-supervised object detection (SSOD) aims to boost the performance of a fully-supervised object detector by exploiting a large amount of unlabeled data. Current state-of-the-art SSOD methods are primarily based on object detectors with many hand-crafted components, e.g., rule-based label assigner~\cite{fastrcnn, yolov3, faster-rcnn,tian2019fcos} and non-maximum suppression (NMS)\cite{nms} post-processing. We term this type of object detector as a traditional object detector. Recently, DETR\cite{detr}, a simple transformer-based end-to-end object detector, has received growing attention. Generally, the DETR-based framework builds upon transformer\cite{transformer} encoder-decoder architecture and generates unique predictions by enforcing a set-based global loss via bipartite matching during training. It eliminates the need for various hand-crafted components, achieving state-of-the-art performance in fully-supervised object detection. Although the performance is desirable, how to design a feasible DETR-based SSOD framework remains under-explored. There are still no systematic ways to fulfill this research gap.

\begin{figure}[ht]
    \centering
    \includegraphics[width=0.8\linewidth]{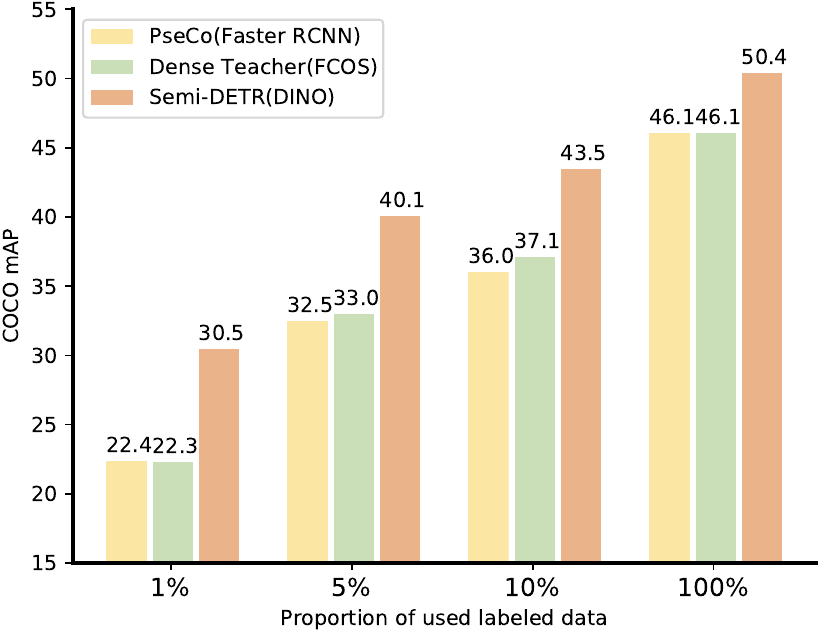}
    \caption{Performance comparisons between the proposed Semi-DETR and other SSOD methods, including PseCo\cite{pseco} and Dense-Teacher\cite{dense-teacher}.}
    \label{fig:comparison}
    \vspace{-6mm}
\end{figure}

Designing an SSOD framework for DETR-based detectors is non-trivial. Concretely, DETR-based detectors take a one-to-one assignment strategy where the bipartite-matching algorithm forces each ground-truth (GT) bounding box to match a candidate proposal as positive, treating remains as negatives. It goes well when the ground-truth bounding boxes are accurate. However, directly integrating DETR-based framework with SSOD is problematic, as illustrated in Fig.~\ref{fig:semi-detr-intro} (a) where a DETR-SSOD vanilla framework utilizes DETR-based detectors to perform pseudo labeling on unlabeled images. In the Teacher-Student architecture, the teacher model usually generates noisy pseudo bounding boxes on the unlabeled images. When the pseudo bounding box is inaccurate, the one-to-one assignment strategy is doomed to match a single inaccurate proposal as positive, leaving all other potential correct proposals as negative, thus causing learning inefficiency. As a comparison, the one-to-many assignment strategy adopted in the traditional object detectors maintains a set of positive proposals, having a higher chance of containing the correct positive proposal. 

On the one hand, the one-to-one assignment strategy enjoys the merits of NMS-free end-to-end detection but suffers the training inefficiency under semi-supervised scenarios; on the other hand, the one-to-many assignment strategy obtains candidate proposal set with better quality making the detector optimized more efficiently but inevitably resulted in duplicate predictions. Designing a DETR-based SSOD framework that embraces these two merits could bring the performance to the next level. 

Additionally, the consistency-based regularization commonly used in current SSOD methods becomes infeasible in DETR-based SSOD. Specifically, current SSOD methods ~\cite{csd,pseco,scale_eq,dsl} utilize consistency-based regularization to help object detectors learn potential feature invariance by imposing consistency constraints on the outputs of pairs-wise inputs (such as scale consistency~\cite{pseco,scale_eq,dsl}, weak-strong consistency\cite{csd}, etc.). Since the input features are deterministic in traditional object detectors, there is a one-to-one correspondence between the inputs and outputs, which makes the consistency constraint convenient to implement. However, this is not the case in DETR-based detectors. DETR-based detectors~\cite{detr,df-detr,dab-detr,dn-detr,dino} use randomly initialized learnable object queries as inputs and constantly update the query features through the attention mechanism. As the query features update, the corresponding prediction results constantly change, which has been verified in \cite{dn-detr}. In other words, there is no deterministic correspondence between the input object queries and its output prediction results, which prevents consistency regularization from being applied to DETR-based detectors. According to the above analysis, we propose a new DETR-based SSOD framework based on the Teacher-Student architecture, which we term Semi-DETR presented in Fig.~\ref{fig:semi-detr-intro} (b). Concretely, we propose a \textbf{Stage-wise Hybrid Matching} module that imposes two stages of training using the one-to-many assignment and the one-to-one assignment, respectively. The first stage aims to improve the training efficiency via the one-to-many assignment strategy and thus provide high-quality pseudo labels for the second stage of one-to-one assignment training. Besides, we introduce a \textbf{Cross-view Query Consistency} module that constructs cross-view object queries to eliminate the requirement of finding deterministic correspondence of object queries and aids the detector in learning semantically invariant characteristics of object queries between two augmented views. Furthermore, we devise a \textbf{Cost-based Pseudo Label Mining} module based on the Gaussian Mixture Model (GMM) that dynamically mines reliable pseudo boxes for consistency learning according to their matching cost distribution. Differently, Semi-DETR is tailored for DETR-based framework, which achieves new SOTA performance compared to the previous best SSOD methods.

To sum up, this paper has the following contributions:
\begin{itemize}[noitemsep, nolistsep]
  \item We present a new DETR-based SSOD method based on the Teacher-Student architecture, called Semi-DETR. To our best knowledge, we are the first to examine the DETR-based detectors on SSOD, and we identify core issues in integrating DETR-based detectors with the SSOD framework.

  \item We propose a stage-wise hybrid matching method that combines the one-to-many assignment and one-to-one assignment strategies to address the training inefficiency caused by the inherent one-to-one assignment within DETR-based detectors when applied to SSOD.  
  
  \item We introduce a consistency-based regularization scheme and a cost-based pseudo-label mining algorithm for DETR-based detectors to help learn semantic feature invariance of object queries from different augmented views.

  \item Extensive experiments show that our Semi-DETR method outperforms all previous state-of-the-art methods by clear margins under various SSOD settings on both MS COCO and Pascal VOC benchmark datasets.
    
\end{itemize}

\section{Related Work}
\textbf{Semi-Supervised Object Detection}. 

In SSOD, Pseudo Labeling~\cite{self-training,stac,ismt,instance-teacher,unbiased-teacher,soft-teacher,debiased-teacher,adap-teacher,double-check} and Consistency-based Regularization~\cite{csd,virtual_consistency,pseco, scale_eq,faster-rcnn,dsl,dense-teacher,unbiased-v2} are two commonly used strategies. A detailed description can be found in the supplementary document. However, most of these works are based on the traditional detectors, e.g. Faster RCNN\cite{faster-rcnn}, which involves many hand-crafted components, e.g anchor box, NMS, etc.  
Our Semi-DETR is significantly different from previous works: (1) we explored the challenges of the DETR-based object detectors on SSOD, which, to our best knowledge, is the first systematic research endeavor in SSOD; (2) our Semi-DETR method is tailored for the DETR-based detectors, which eliminates the training efficiency caused by bipartite matching with the noisy pseudo labels and presents a new consistency scheme for set-based detectors.

\textbf{End-to-End Object Detection with Transformer}. 
The pioneering work DETR\cite{detr} introduced transformers into object detection to eliminate the need for complex hand-crafted components in traditional object detectors. Many follow-up works have been dedicated to solving the slow convergence and high complexity issues of DETR~\cite{df-detr,c-detr,dab-detr,dn-detr,dino,efficient}. Recently, DINO~\cite{dino} combined with a variety of improvements related to DETR, such as query selection\cite{df-detr,efficient}, contrastive query denoising\cite{dn-detr}, and achieved SOTA performance across various object detection benchmark datasets with excellent convergence speed. Complementary to these, we aim to extend the study of DETR-based detectors to SSOD and present Semi-DETR, which is a tailored design for SSOD. Our framework is agnostic to the choice of DETR-based detectors and could easily integrate with all DETR-based detectors. Omni-DETR~\cite{omni} is a DETR-based object detector designed for omni-supervised object detection. It is not designed specifically for SSOD as admitted in their paper, but it is extended to the SSOD task by introducing a simple pseudo-label filtering scheme. Our Semi-DETR is significantly different from Omni-DETR in the following aspects: (1) Different motivations for model design; (2) Different training strategy; (3) Significant performance improvement. The detailed discussion is in Supplementary Document.

\begin{figure*}[ht]
    \centering
    \includegraphics[width=0.8\linewidth]{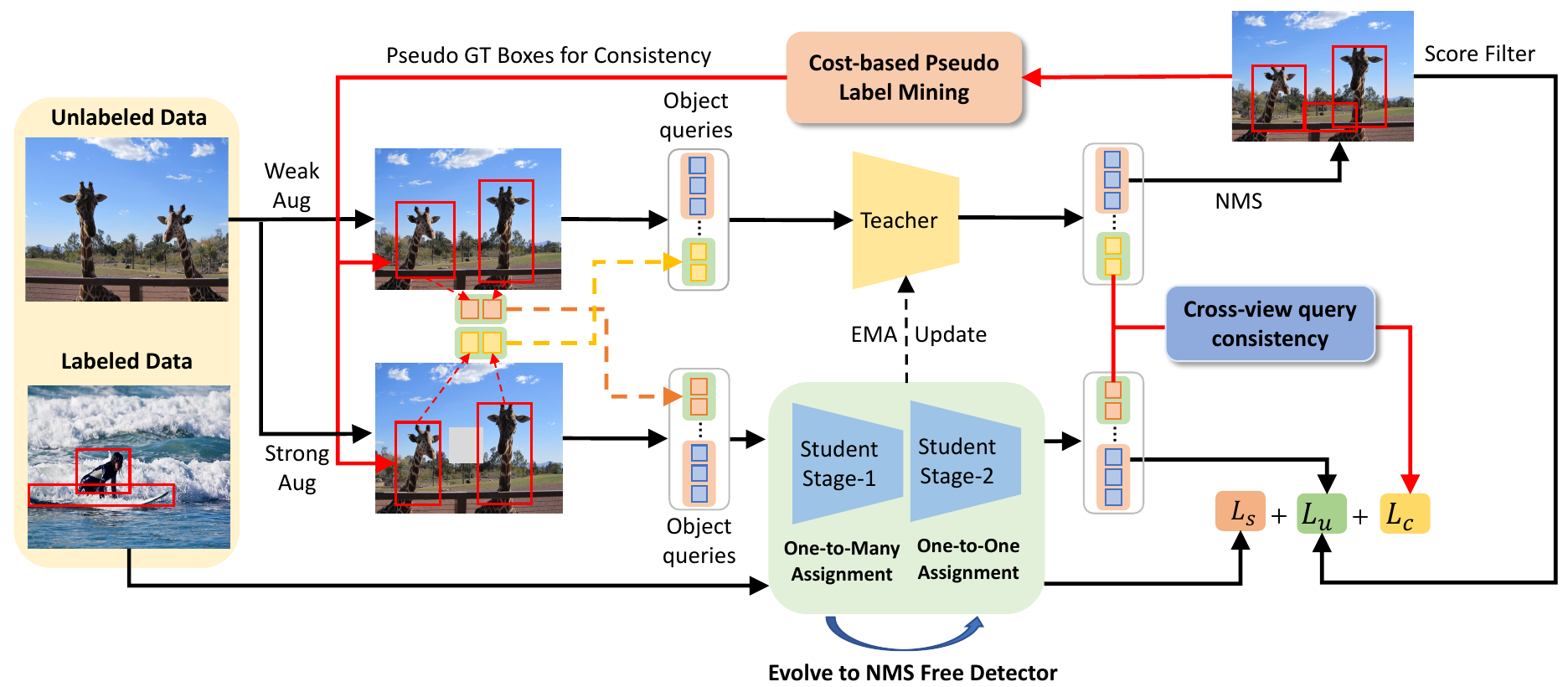}
    \caption{Overview of Semi-DETR. Our framework is based on Teacher-Student architecture. Specifically, a multi-stage training strategy is derived to avoid incorrect bipartite matching with low-quality pseudo labels. Hybrid Matching with one-to-many assignment is applied in the first stage to generate higher quality pseudo labels for the following one-to-one training stage. Besides, a cross-view query consistency loss is designed to further enhance the consistency learning for the whole training process, where the pseudo boxes are filtered by a cost-based GMM mining module.}
    
    \label{fig:network}
    \vspace{-3mm}
\end{figure*}

\section{Semi-DETR}

\subsection{Preliminary}
We aim to address the problem of semi-supervised DETR-based object detection, where a labeled image set $D_s =\{x_i^s, y_i^s\}_{i=1}^{N_s}$ and an unlabeled image set $D_u=\{x_i^u\}_{i=1}^{N_u}$ are available during training. $N_s$ and $N_u$ denote the amount of labeled and unlabeled images. For the labeled images $x^s$, the annotations $y^s$ contain the coordinates and object categories of all bounding boxes. 

\subsection{Overview}
The overall framework of our proposed Semi-DETR is illustrated in Fig.~\ref{fig:network}. Following the popular teacher-student paradigm~\cite{tarvainen2017mean} for SSOD, our proposed Semi-DETR adopts a pair of teacher and student models with exactly the same network architecture. Here we adopt DINO~\cite{dino} as an example while the overall framework of Semi-DETR is compatible with other DETR-based detectors. Specifically, in each training iteration, weak-augmented and strong-augmented unlabeled images are fed to the teacher and student, respectively. Then the pseudo labels generated by the teacher with confidence scores larger than $\tau_s$ served as supervisions for training the student. The parameters of the student are updated by back-propagation, while the parameters of the teacher model are the exponential moving average (EMA) of the student. 

Our main contribution contains three new components: stage-wise hybrid matching, cross-view query consistency, and cost-based pseudo-label mining, which address the core issues 
of DETR-based SSOD. In the following sections, we introduce more details of our proposed Semi-DETR.

\subsection{Stage-wise Hybrid Matching}

DETR-based frameworks rely on one-to-one assignment for end-to-end object detection. For DETR-based SSOD framework, an optimal one-to-one assignment $\hat{\sigma}_{o2o}$ can be obtained by performing the Hungarian algorithm between the predictions of the student and pseudo-labels generated by the teacher:
\begin{equation}
  \hat{\sigma}_{o2o} = \underset{\sigma \in \xi_{N}}{\arg \min } \sum_{i=1}^{N} \mathcal{C}_{\operatorname{match}}\left(\hat{y}^{t}_{i}, \hat{y}^{s}_{\sigma(i)}\right)
  \label{eq:one2one}
\end{equation}
where $\xi_{N}$ is the set of permutations of $N$ elements and $\mathcal{C}_{\operatorname{match}}\left(\hat{y}^{t}_{i}, \hat{y}^{s}_{\sigma(i)}\right)$ is the matching cost between the pseudo-labels $\hat{y}^t$ and the prediction of the student with index $\sigma(i)$. 

However, in the early stage of SSOD training, the pseudo-labels generated by the teacher are usually inaccurate and unreliable, which imposes a high risk of generating sparse and low-quality proposals under the one-to-one assignment strategy. To exploit multiple positive queries to realize efficient semi-supervised learning, we propose to replace the one-to-one assignment with the one-to-many assignment:
\begin{equation}
  \hat{\sigma}_{o2m} =\left\{\underset{\bm{\sigma_i} \in C^M_{N}}{\arg \min } \sum_{j=1}^{M} \mathcal{C}_{\operatorname{match}}\left(\hat{y}^{t}_{i}, \hat{y}^{s}_{\bm{\sigma_i(j)}}\right)\right\}_{i=1}^{|\hat{y}^t|}.
  \label{eq:one2many}
\end{equation}
where $C^M_{N}$ is the combination of $M$ and $N$, which denotes that a subset of $M$ proposals is assigned to each pseudo box $\hat{y}^t_{i}$. Following\cite{ota,tood}, we utilize a high-order combination of classification score $s$ and the IoU value $u$  as the matching cost metric:
\begin{equation}
  m =  s^{\alpha}\cdot u^{\beta}
  \label{eq:match_criterion}
\end{equation}
where $\alpha$ and $\beta$ control the effect of classification score and IoU during the assignment, and following \cite{tood}, we set $\alpha=1$, $\beta=6$ by default. With the one-to-many assignment, $M$ proposals with the largest $m$ values are selected as positive samples while regarding the remaining proposals as negative ones. \par

We train the model with one-to-many assignment for $T_1$ iterations in the early stage of semi-supervised training. Following \cite{tood,gfl}, the classification loss and regression loss are also modified at this stage:
\begin{equation}
  \mathcal{L}^{o2m}_{c l s}=\sum_{i=1}^{N_{\text {pos }}}\left|\hat{m}_{i}-s_{i}\right|^{\gamma} B C E\left(s_{i}, \hat{m}_{i}\right)+\sum_{j=1}^{N_{\text {neg }}} s_{j}^{\gamma} B C E\left(s_{j}, 0\right)
    \label{eq:o2m_cls_loss}
\end{equation}
\begin{equation}
 \mathcal{L}^{o2m}_{r e g}=\sum_{i=1}^{N_{\text {pos }}} \hat{m}_{i} \mathcal{L}_{G I o U}\left(b_{i}, \hat{b}_{i}\right)+\sum_{i=1}^{N_{\text {pos }}} \hat{m}_{i} \mathcal{L}_{L_{1}}\left(b_{i}, \hat{b}_{i}\right)
  \label{eq:o2m_reg_loss}
\end{equation}
\begin{equation}
\mathcal{L}^{o2m} = \mathcal{L}^{o2m}_{cls} + \mathcal{L}^{o2m}_{reg}
\end{equation}
where $\gamma$ is set to 2 by default. With multiple assigned positive proposals for each pseudo label, the potentially high-quality positive proposals also get the chance to be optimized, which greatly improves the convergence speed and, in turn, obtains pseudo labels with better quality. However, the multiple positive proposals for each pseudo label result in duplicate predictions. To mitigate this problem, we propose to switch back to the one-to-one assignment training in the second stage. By doing this, we enjoy the high-quality pseudo labels after the first stage training and gradually reduce duplicate predictions to reach an NMS-free detector with one-to-one assignment training at the second stage. The loss functions of this stage are the same as \cite{dino}:
\begin{equation}
\mathcal{L}^{o2o} = \mathcal{L}^{o2o}_{cls} + \mathcal{L}^{o2o}_{reg}
\end{equation}

\begin{figure}
    \centering
    \includegraphics[width=\linewidth]{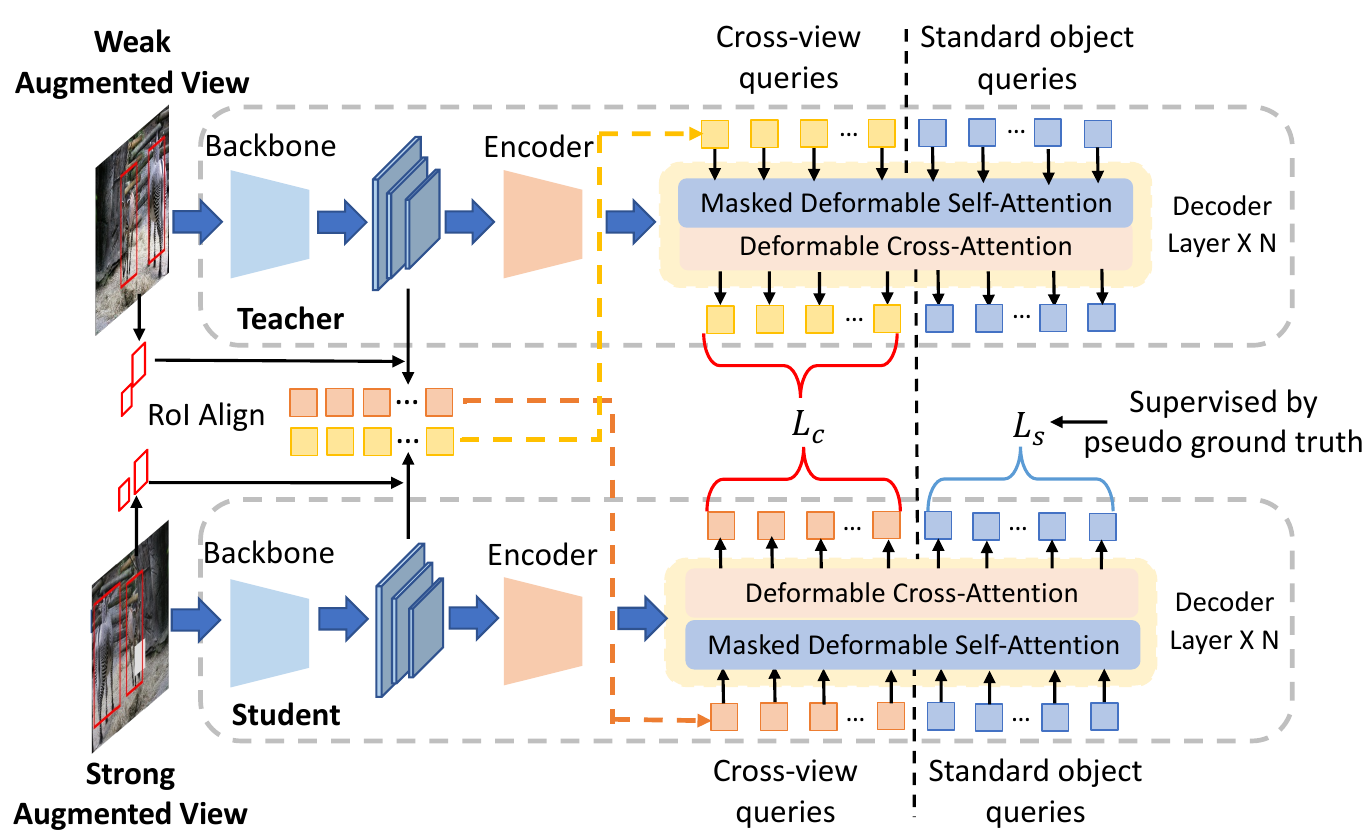}
    \caption{Overview of the cross-view query consistency module. Query embeddings from the RoI features of pseudo labels on different views are cross-wise sent to the teacher and student decoders. The corresponding decoded features are enforced to be similar by a consistency loss.}
    \label{fig:consistency_pipeline}
    \vspace{-3mm}
\end{figure}

\subsection{Cross-view Query Consistency}
Traditionally, in non-DETR-based SSOD frameworks, consistency regularization can be employed conveniently by minimizing the difference between the output of teacher $f_{\theta}$ and student $f_{\theta}^{\prime}$, given the same input $x$ with different stochastic augmentation:
\begin{equation}
        \mathcal{L}_{c} =\sum_{\mathbf{x} \in \mathcal{D}_{u}} \operatorname{MSE}\left(f_{\theta}(\mathbf{x}), f_{\theta}^{\prime}(\mathbf{x})\right)
\end{equation} 
However, for DETR-based frameworks, as there is no clear (or deterministic) correspondence between the input object queries and their output prediction results, conducting consistency regularization becomes infeasible. To overcome this issue, we propose a Cross-view Query Consistency module that enables the DETR-based framework to learn semantically invariant characteristics of object queries between different augmented views.

Fig.~\ref{fig:consistency_pipeline} illustrates our proposed cross-view query consistency module. Specifically, for each unlabeled image, given a set of pseudo bounding boxes $b$, we process the RoI features extracted via RoIAlign\cite{roi-align} with several MLPs:
\begin{equation}
\begin{aligned}
   c_t = \operatorname{MLP}(\operatorname{RoIAlign}(F_t, b)) \\
   c_s = \operatorname{MLP}(\operatorname{RoIAlign}(F_s, b)) 
    \label{eq:roi_align}
\end{aligned}
\end{equation}
where $F_t$ and $F_s$ denote the backbone feature of the teacher and student, respectively.
Subsequently, $c_t$ and $c_s$ are regarded as cross-view query embeddings and attached to the original object queries in another view to serve as the input of the decoder:
\begin{equation}
\begin{aligned}
    \hat{o}_{t},o_{t} &= \operatorname{Decoder}_t([c_s,q_t], E_t| A) \\
    \hat{o}_{s},o_{s} &= \operatorname{Decoder}_s([c_t,q_s], E_s| A)
\end{aligned}
    \label{eq:cross_view decoding}
\end{equation}
where ${q_{\cdot}}$ and $E_{\cdot}$ denote the original object queries and the encoded image features, respectively. $\hat{o}_{\cdot}$ and ${o_{\cdot}}$ denote the decoded features of cross-view queries and original object queries. Note the subscript $t$ and $s$ indicate teacher and student, respectively. Following \cite{dn-detr}, the attention mask $A$ is also employed to avoid information leakage.

With the semantic guide of input cross-view queries embeddings, the correspondence of the decoded features can be naturally guaranteed, and we impose consistency loss as follows:
\begin{equation}
        \mathcal{L}_{c} =\operatorname{MSE}(\hat{o}_{s}, \operatorname{detach}(\hat{o}_{t}))
\end{equation}

\subsection{Cost-based Pseudo Label Mining}
To mine more pseudo boxes with meaningful semantic contents for the cross-view query consistency learning, we propose a cost-based pseudo label mining module that dynamically mines reliable pseudo boxes in the unlabeled data. Specifically, we perform an additional bipartite matching between the initial filtered pseudo boxes and the predicted proposals and utilize the matching cost to describe the reliability of the pseudo boxes:
\begin{equation}
  C_{ij} = \lambda_1C_{Cls}(p_i,\hat{p_j}) + \lambda_2C_{GIoU}(b_i,\hat{ b_j}) + \lambda_3C_{L_1}(b_i,\hat{b_j})
  \label{eq:match_cost}
\end{equation}
where $p_i$, $b_i$ represents the classification and regression result of $i$-th predicted proposals while $\hat{p_j}$, $\hat{b_j}$ indicates the class label and box coordinates of $j$-th pseudo label.

\begin{figure*}[htbp]
  \centering
  \begin{subfigure}{0.22\linewidth}
    \centering
    \includegraphics[width=4cm,height=3cm]{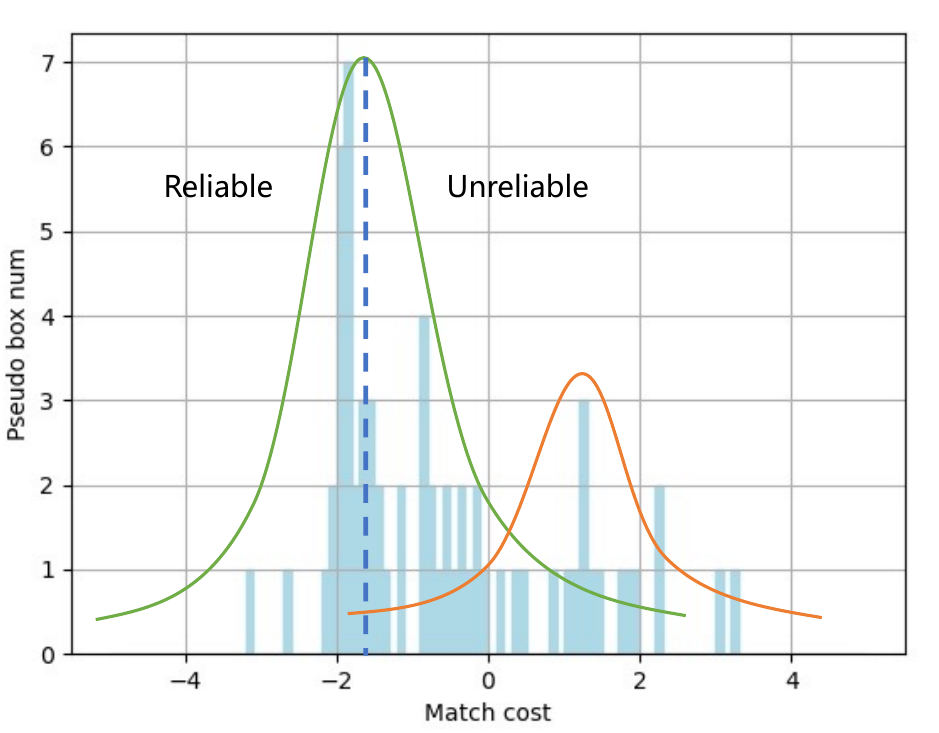}
    \caption{Distribution of Cost}
    \label{a}
  \end{subfigure}
  \centering
  \begin{subfigure}{0.22\linewidth}
    \centering
    \includegraphics[width=4cm,height=3cm]{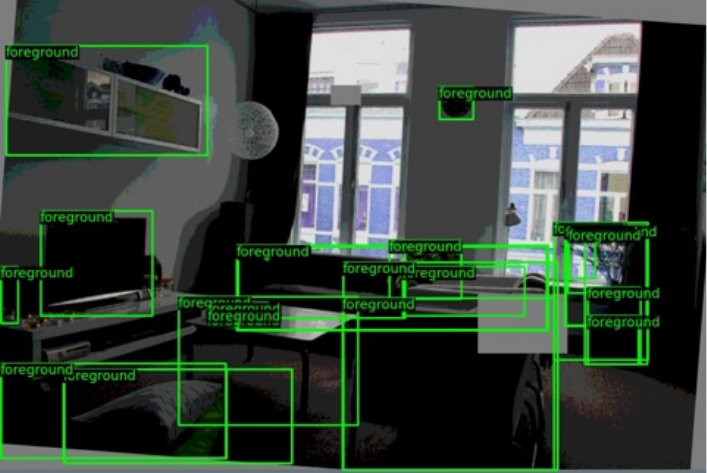}
    \caption{Initial Pseudo Boxes}
    \label{b}
  \end{subfigure}
  \centering
  \begin{subfigure}{0.22\linewidth}
    \centering
    \includegraphics[width=4cm,height=3cm]{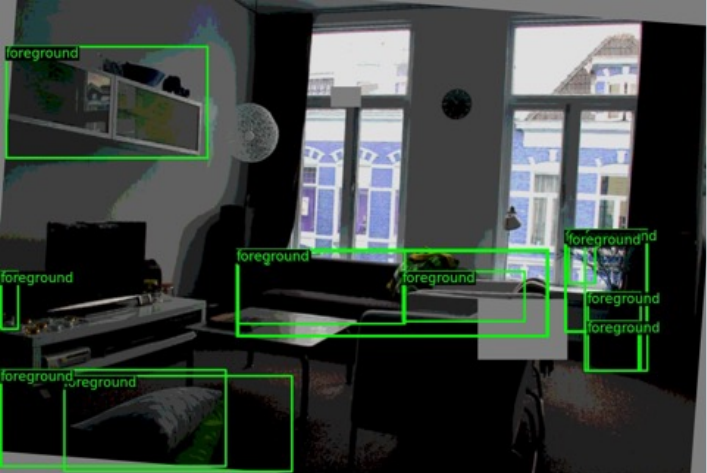}
    \caption{Unreliable Pseudo Boxes}
  \label{c}
  \end{subfigure}
  \begin{subfigure}{0.22\linewidth}
    \centering
    \includegraphics[width=4cm,height=3cm]{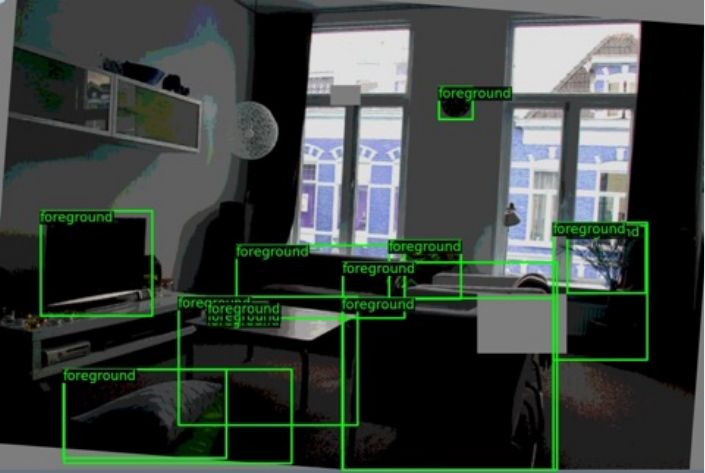}
    \caption{Reliable Pseudo Boxes}
    \label{d}
  \end{subfigure}
    \caption{ An illustration of Cost-based Pseudo Label Mining. We first take the image-level confidence score's mean with variance to get the initial pseudo labels, shown in (b), for each image and perform the Hungarian match to get the matching cost of each pseudo ground truth bounding box within a batch. Then, we fit a GMM model with these cost values, shown in (a). We argue that the pseudo boxes with lower cost are more likely to be the reliable pseudo boxes, so we take the lower threshold from the GMM model to filter the pseudo label again to obtain the final pseudo boxes presented in (d).}
  \label{cost_gmm}
  \vspace{-5mm}

\end{figure*}

Subsequently, in each training batch, we cluster the initial pseudo boxes into two states by fitting a Gaussian Mixture Model for the matching cost distribution. As illustrated in Fig.~\ref{cost_gmm}, the matching cost aligns well with the quality of pseudo boxes. We further set the cost value of the clustering center of the reliable ones as the threshold and collect all pseudo boxes with lower cost than the threshold for the cross-view query consistency calculation.

\subsection{Loss Function}
The final loss $L$ is represented as follows:

\begin{equation}
\begin{aligned}
    \mathcal{L} &= \mathbb{I}(t\leq T_1)\cdot(\mathcal{L}^{o2m}_{sup}+w_u\cdot \mathcal{L}^{o2m}_{unsup}) \\
    &+  \mathbb{I}(t > T_1)\cdot(\mathcal{L}^{o2o}_{sup}+w_u\cdot \mathcal{L}^{o2o}_{unsup}) \\
    &+  w_c\cdot \mathcal{L}_c
\end{aligned}
\end{equation}
where $\mathcal{L}^{\cdot}_{sup}$ and $\mathcal{L}^{\cdot}_{unsup}$ are the supervised loss and the unsupervised loss, respectively, containing both the classification loss and regression loss. The $\mathcal{L}_c$ means the cross-view consistency loss. The $w_u$ and $w_c$ are the unsupervised loss weight and consistency loss weight, which set $w_u=4$ and $w_c=1$ by default. $t$ is the current training iteration and $T_1$ is the duration time of the first stage training within the SHM module.

\section{Experiments}
\label{sec:gmm_pseudo_labels}

\subsection{Datasets and Evaluation Metrics}
We validate our method on the MS-COCO benchmark\cite{coco} and Pascal VOC datasets\cite{voc}. MS-COCO contains 80 classes with 118k labeled images in the train2017 set and 123k unlabeled images in the unlabeled2017 set. In addition, the val2017 set with 5k images is provided for validation. Following \cite{soft-teacher}, we consider two evaluation settings to validate our method on the MS-COCO benchmark: (1) \textbf{COCO-Partial}. 1\%, 5\%, and 10\% images of the COCO train2017 set are randomly sampled as the labeled training data, and the remaining images of train2017 are regarded as the unlabeled data. 5 different data folds are created for each data split to validate our method. The average of standard COCO mAP on the val2017 is adopted as our final performance metric. (2) \textbf{COCO-Full}. Under this setting, the entire train2017 is utilized as the labeled data, and unlabeled2017 is used as the additional unlabeled data. The standard COCO mAP on the val2017 is taken as the evaluation metric. \textbf{Pascal VOC} contains 20 classes with VOC2007 and VOC2012 provided as the labeled data and unlabeled data respectively. The evaluation metrics are the COCO-style $AP_{50:95}$ and $AP_{50}$ on the VOC2007 test set. 

\subsection{Implementation Details}
To avoid loss of generality, we choose Deformable DETR\cite{df-detr} and DINO to integrate into our Semi-DETR method. Following them, we use ResNet-50\cite{resnet} pre-trained on ImageNet\cite{imagenet} as our backbone network. Focal Loss\cite{retinanet} is used for classification during training. Smooth L1 Loss and GIoU\cite{giou} Loss are used for regression. We set the number of object queries to 300 for Deformable DETR and 900 for DINO, respectively. For the training hyperparameters, following \cite{soft-teacher}: (1) For the COCO-Partial benchmark, we train Semi-DETR for 120k iterations on 8 GPUs with 5 images per GPU. The first stage with one-to-many assignment is set to 60k iterations. The ratio of the labeled data and unlabeled data is set to 1:4. The weight of the unsupervised loss is set to $\alpha = 4.0$. (2) For the COCO-Full benchmark, we double the training time with COCO-unlabeled to 240k, where the first stage with one-to-many assignment is set to 180k iterations. The batch size is set to 64 on 8 GPUs with 8 images per GPU. The ratios of labeled data and unlabeled data are set to 1:1, and the loss weight of unlabeled data is set to $\alpha = 2.0$. (3) For the Pascal VOC benchmark, we train Semi-DETR for 60k iterations where The first stage with one-to-many assignment is set to 40k iterations. Other settings are kept the same with COCO-Partial. For all experiments, the confidence threshold is set to 0.4. We utilize Adam\cite{adam} with a learning rate of 0.001, and no learning rate decay scheme is used. The teacher model is updated through EMA with a momentum of 0.999. Besides, we follow the same data prepossessing, and augmentation pipeline in \cite{soft-teacher} without modifications.


\begin{table*}[!htbp]
\centering
\caption{Comparisons with SOTA SSOD methods under the COCO-Partial setting. All results are the average of all 5 folds. Def-DETR denotes Deformable DETR. Sup Only denotes supervised only baseline.}
\begin{tabular}{c|c|c||c|c|c}
\hline  
Category&Method&ID & 1\% & 5\% & 10\% \\ 
\hline 
 &Unbiased Teacher&1 &20.75 ± 0.12&28.27 ± 0.11&31.50 ± 0.10\\
Two-Stage&Soft-Teacher&2 & 20.46 ± 0.39& 30.74 ± 0.08&34.04 ± 0.14\\
&PseCo&3 & 22.43 ± 0.36& 32.50 ± 0.08&36.06 ± 0.24\\
\hline
& DSL&4 & 22.03 ± 0.28& 30.87 ± 0.24&36.22 ± 0.18\\
One-Stage&Dense Teacher&5 & 22.38 ± 0.31& 33.01 ± 0.14&37.13 ± 0.12\\
& Unbiased Teacher v2&6 & 22.71 ± 0.42& 30.08 ± 0.04 &32.61 ± 0.03\\
\hline  
 & Omi-DETR(Def-DETR) & 7 &18.60 &30.20 & 34.10 \\
& Def-DETR(Sup only)& 8 &11.00 ± 0.24 &23.70 ± 0.13&29.20 ± 0.11 \\
& Def-DETR SSOD(Baseline)&9 &19.40 ± 0.31 &31.10 ± 0.21&34.80 ± 0.09 \\
& Semi-DETR(Def-DETR)&10 &\textbf{25.20 ± 0.23} &\textbf{34.50 ± 0.18}&\textbf{38.10 ± 0.14} \\
End-to-End & DINO(Sup only)&11 &18.00 ± 0.21    &29.50 ± 0.16&35.00 ± 0.12 \\
&DINO SSOD (Baseline)&12 & 28.40 ± 0.21 & 38.00 ± 0.13&41.60 ± 0.11\\
 & Omi-DETR(DINO) & 13 &27.60 &37.70 &41.30 \\
&Semi-DETR(DINO)&14 & \textbf{30.50 ± 0.30}&\textbf{40.10 ± 0.15} &\textbf{43.50 ± 0.10}\\
\hline 
\end{tabular}
\vspace{-3mm}
\label{table_coco_partial}
\end{table*}

\begin{table}[!h]
\centering
\caption{Comparisons with SOTA SSOD methods under the Pascal VOC setting. Def-DETR denotes Deformable DETR. Sup Only denotes supervised only baseline.}
\vspace{-3mm}
\resizebox{\linewidth}{!}{
\begin{tabular}{c|c|ccc}
\hline 
Category&Method & $AP_{50}$&$AP_{50:95}$ \\
\hline 
 &Unbiased Teacher &77.37&48.69\\

Two-Stage&Soft-Teacher &-&-\\

&PseCo  &-&-\\

\hline
& DSL &80.70&56.80\\
One-Stage& Dense Teacher &79.89&55.87\\
& Unbiased Teacher v2 &81.29&56.87\\
\hline  
 & Def-DETR(Sup only) &74.50 &46.20\\
  & Def-DETR SSOD(Baseline) &78.90 &53.40\\
   & Semi-DETR(Def-DETR) &\textbf{83.50} &\textbf{57.20}\\
 & DINO(sup only) &81.20 &59.60\\

End-to-End & DINO SSOD (Baseline) &84.30&62.20\\
&Semi-DETR(DINO)  &\textbf{86.10}&\textbf{65.20}\\
\hline 
\end{tabular}}
\label{table_pascal_voc}
\vspace{-3mm}
\end{table}

\subsection{Comparison with SOTA methods}
We compare our Semi-DETR method with current SOTA SSOD methods on both MS-COCO and Pascal VOC datasets. We present the superiority of Semi-DETR in the following aspects: (1) comparisons to two-stage and one-stage detectors, (2) comparisons to DETR-based detectors, and (3) generalization ability.

\textbf{COCO-Partial benchmark}. According to Tab.~\ref{table_coco_partial}, Semi-DETR shows significant superiority over current SOTA SSOD methods across all experiment settings in COCO-Partial. Concretely, (1) compared to SOTA two-stage and one-stage detectors, Semi-DETR outperforms PseCo (experiment 3) by 2.77, 2.00, 2.05 mAP with Deformable DETR (by 8.07, 7.60, 7.44 mAP with DINO) under the 1\%, 5\%, 10\% settings and beats Dense Teacher (experiment 5) by 2.82, 1.49, 0.97 mAP with Deformable DETR (by 8.12, 7.09, 6.37 mAP with DINO) under the 1\%, 5\%, 10\% settings. Obviously, Semi-DETR is a better semi-supervised object detector, and it does not require hand-crafted components used in two-stage and one-stage detectors; (2) we construct two DETR-based baselines, namely DETR under supervised training only and a simple pseudo labeling Teacher-Student architecture integrating DETR with SSOD. By comparing experiments 7-10 (or experiments 11-14), Semi-DETR outperforms the supervised baseline by 14.20, 10.80, 8.90 mAP with Deformable DETR (12.50, 10.60, 8.50 mAP with DINO) and surpasses the SSOD baseline by 5.80, 3.40, 3.30 mAP with Deformable DETR (2.10, 2.10, 1.90 mAP with DINO). This demonstrates that simply integrating DETR-based detectors with Teacher-Student architecture is not optimal. 

(3) we use Deformable DETR and DINO to show the generalization ability of our Semi-DETR method. Apparently, Semi-DETR consistently boosts the performance of both detectors over the corresponding baselines by clear margins (experiments 7-14). With stronger detectors like DINO, Semi-DETR still enjoys a notable performance improvement. 

\textbf{COCO-Full benchmark}. According to Tab.~\ref{table_coco_full}, when adding additional unlabeled2017 data, Semi-DETR with Deformable DETR enjoys 3.6 mAP performance gain and reaches 47.2 mAP, surpassing PseCo and Dense Teacher by 1.1 and 1.1 mAP, respectively. This further manifests the effectiveness of Semi-DETR. Besides, under stronger baselines like DINO, Semi-DETR still shows obvious performance gain (+1.8 mAP), which outperforms PseCo and Dense Teacher by 4.3 and 4.3 mAP respectively, and generates a new SOTA performance of 50.4 mAP. 

\textbf{Pascal VOC benchmark}. Semi-DETR presents consistent performance improvements on the Pascal VOC benchmark as shown in Tab.~\ref{table_pascal_voc}. Generally, Semi-DETR outperforms the supervised baseline by 9.0 on $AP_{50}$ and 11.0 on $AP_{50:95}$ with Deformable DETR (by 4.9 on $AP_{50}$ and 5.6 on $AP_{50:95}$ with DINO). Furthermore, Semi-DETR beats all previous SOTA SSOD methods by significant margins on both $AP_{50}$ and $AP_{50:95}$.

\begin{table}
  \centering
  \caption{Comparisons with SOTA SSOD methods under the COCO-Full setting.  Def-DETR denotes Deformable DETR. Sup Only denotes supervised only baseline.}
  \vspace{-3mm}
  \begin{tabular}{cccccc}
    \toprule
    Method & 100\% \\
 
    \hline
    Unbiased Teacher & 40.2 $\stackrel{+1.1}{\longrightarrow}$ 41.3 \\
    Soft-Teacher & 40.9 $\stackrel{+3.6}{\longrightarrow}$ 44.5 \\
    PseCo & 41.0 $\stackrel{+5.1}{\longrightarrow}$ 46.1 \\
    \hline
    DSL & 40.2 $\stackrel{+3.6}{\longrightarrow}$ 43.8 \\
    Dense Teacher & 41.2 $\stackrel{+3.6}{\longrightarrow}$ 46.1 \\
    \hline
     Semi-DETR(Def-DETR) & \textbf{43.6} $\stackrel{+3.6}{\longrightarrow}$ \textbf{47.2} \\
    Semi-DETR(DINO) & \textbf{48.6} $\stackrel{+1.8}{\longrightarrow}$ \textbf{50.4} \\
    \bottomrule
  \end{tabular}
  \vspace{-3mm}
  \label{table_coco_full}
\end{table}

\subsection{Ablation Study}
We conduct extensive experiments to verify the effectiveness of Semi-DETR in the following aspects: (1) component effectiveness; (2) variants of Stage-wise Hybrid Matching (SHM);  
(3) effectiveness of Cross-view Query Consistency (CQC) and Cost-based Pseudo Label Mining (CPM); (4) hyper-parameters. All experiments are performed with DINO as the base detector on the 10\% labeled images setting of the COCO-Partial benchmark. 

\textbf{Component Effectiveness}. According to Tab.~\ref{table_ablation}, we perform four experiments to verify the effectiveness of each proposed component. We formulate a strong baseline that integrates DINO with SSOD via pseudo labeling in experiment 1. In general, our proposed components enjoy consistent performance improvements. Specifically, by introducing the SHM module, it outperforms the baseline by 1.1 mAP. Further integrating the CQC and CPM modules brings an extra 0.8 improvement. This shows that our proposed components are complementary to each other and proves the effectiveness of each component in our model.

\begin{table}[h]
\centering
\caption{Component effectiveness of Semi-DETR. SHM denotes the Stage-wise Hybrid Matching, CQC means Cross-view Query Consistency, and CPM represents Cost-based Pseudo Label Mining, respectively.}
\vspace{-3mm}
\begin{tabular}{c||c|c|c|c|c|c}
\hline  
ID& SHM& CQC & CPM&mAP&$AP_{50}$&$AP_{75}$\\
\hline  
1&  & & & 41.6&58.3&45.1\\
2& \checkmark & & &42.7&59.3&46.2\\
3& \checkmark& \checkmark&& 43.1&59.6&46.6\\
4& \checkmark & \checkmark&\checkmark&\textbf{43.5}&\textbf{59.7}&\textbf{46.8}\\
\hline 
\end{tabular}
\label{table_ablation}
\end{table}

\textbf{Variants of SHM}.
We examine the impact of different one-to-many assignment strategies within SHM in the first stage of training. Concretely, Max-IoU\cite{faster-rcnn}, ATSS\cite{atss} and SimOTA\cite{yolox} are chosen as the alternatives. All models are trained for 60k iterations. As presented in Tab.~\ref{ablation_assignment},

it is interesting to find that not all traditional one-to-many assignment methods are effective in DETR-based detectors. Max-IoU assignment strategy and ATSS show significant performance degradation when applied to the first stage, even though they are commonly used in traditional object detectors. On the other hand, SimOTA shows comparable performance to our one-to-many assignment strategy. 

This is possibly caused by the fact that SimOTA and our method adopt a ranking-based one-to-many assignment strategy while Max-IoU and ATSS utilize hard or dynamic thresholding-based one-to-many assignment strategy, which leads to a significant difference number of assigned positive samples for each pseudo ground truth bounding box and thus suffers performance degradation. More analysis can be found in the supplementary document.

\textbf{Effectiveness of CQC+CPM}. 
According to Tab.~\ref{ablation_pseudo_method}, we compare four different methods to generate pseudo labels for CQC and evaluate the precision and recall metrics of the generated pseudo labels. First, we present two methods (by setting a fixed classification score $\tau_s = 0.4$ or by selecting Top-K pseudo labels with the highest confidence scores) that obtain pseudo labels with high precision (81.5\% or 80.2\%) and low recall (41.3\% or 39.4\%) but observe marginal performance gains. Then we present the Mean+Std method that aims to balance the precision (60.2\%) and recall (54.0\%) of pseudo labels via combining the image-level mean confidence score and variance $\tau = \mu + \sigma$, which enjoys a better performance improvement (+0.4 mAP). Finally, our Cost-based GMM method achieves a better trade-off between the precision (77.6\%) and recall (52.1\%) metrics, which has a 0.8 performance gain.

\begin{table}[h]
\centering
\caption{Effects of different methods to filter pseudo labels for cross-view consistency training.}
\vspace{-3mm}
\begin{tabular}{c|c|c|c}
\hline  
Method& mAP& Precision & Recall\\
\hline  
Fixed(0.4) &42.8&81.5\%& 41.3\%\\
Top-K(K=9) &42.9&80.2\%&39.4\%\\
Mean + Std & 43.1&60.2\%&54.0\%\\
Cost-based GMM & 43.5&77.6\%&52.1\%\\
\hline 
\end{tabular}
\label{ablation_pseudo_method}
\vspace{-3mm}
\end{table}

\begin{table}[h]
\centering
\caption{Effects of the different one-to-many assignment methods in the first stage.}
\vspace{-3mm}
\begin{tabular}{c|c|c|c}
\hline  
Strategy& mAP&$AP_{50}$ & $AP_{75}$\\
\hline  
Max-IoU &11.4&15.0&12.1\\
ATSS &18.7&30.5&18.9\\
SimOTA & 42.5& \textbf{59.9}&45.2\\
Ours &\textbf{ 42.8}&59.8&\textbf{46.0}\\
\hline 
\end{tabular}
\label{ablation_assignment}
\vspace{-3mm}
\end{table}

\begin{table}[h]
\centering
\caption{Effects of the training iteration $T_1$ of the first stage using one-to-many assignment strategy in Stage-wise Hybrid Matching.}
\vspace{-3mm}
\begin{tabular}{cccccc}
\toprule  
$T_1$& 40k&  60k&80k&100k&120k\\
\midrule  
mAP &42.9&\textbf{43.5}&43.2 &43.0&\textbf{44.0} \\
NMS-Free& Y & Y & Y & Y & N\\

\bottomrule 
\end{tabular}
\vspace{-3mm}
\label{ablation_warm_up_iteration}
\end{table}

\begin{table}[h]
\centering
\caption{Effects of the pseudo label threshold $\tau_s$.}
\vspace{-3mm}
\begin{tabular}{c|ccccc}
\hline  
$\tau_s$ & 0.2&0.3&0.4&0.5&0.6\\
\hline  
mAP & 42.6& 43.0&\textbf{43.5}&43.2&42.8\\
\hline 
\end{tabular}
\vspace{-7mm}
\label{table_unsup_threshold}
\end{table}

\textbf{Hyperparameters.} 
We study two types of hyperparameters in our model: (1) the pseudo label threshold $\tau_s$; (2) the training iterations $T_1$ of the first stage in SHM. For $\tau_s$, according to Tab.~\ref{table_unsup_threshold}, the best performance is achieved when $\tau_s=0.4$. Possibly, a lower threshold could introduce noisy pseudo labels, while a higher threshold could decrease the effective number of pseudo labels. For the training iterations of the first stage, according to Tab.~\ref{ablation_warm_up_iteration}, performing the one-to-many assignment strategy across both stages achieves 44.0 mAP at the cost of using NMS in the end. The appropriate training time of the first stage is at 60k iterations, which achieves the best performance of 43.5 mAP and does not require NMS post-process at the same time. 
\section{Conclusion}
We analyzed the challenges of the DETR-based object detectors on semi-supervised object detection, including the learning inefficiency of one-to-one assignment with inaccurate pseudo labels and the difficulties of designing consistency-based regularization due to the absence of deterministic correspondence from object queries. We proposed Semi-DETR, the first transformer-based end-to-end semi-supervised object detector. It consists of a Stage-wise Hybrid Matching method that embraces the merits of both one-to-many assignment and one-to-one assignment strategies, a Cross-view Query Consistency method that learns semantic feature invariance of object queries from different views via unlabeled images, and a Cost-based Pseudo Labeling module that adaptively mines more reliable pseudo labels for improving the efficiency of consistency training. 
Extensive experiments demonstrate the superiority of Semi-DETR on both MS-COCO and Pascal VOC benchmarks. 

\section*{Acknowledgments}
This work was supported in part by the Guangdong Basic and Applied Basic Research Foundation (NO.~2020B1515020048), in part by the National Natural Science Foundation of China (NO.~61976250), in part by the Shenzhen Science and Technology Program (NO.~JCYJ20220530141211024) and in part by the Fundamental Research Funds for the Central Universities under Grant 22lgqb25.

{\small
\bibliographystyle{ieee_fullname}
\bibliography{egbib}
}

\newpage

\renewcommand\thefigure{\thesection.\arabic{figure}}
\renewcommand\thetable{\thesection.\arabic{table}}

\setcounter{page}{1}
\begin{center}
\section*{Supplementary Material}
\end{center}

\section*{Extended Details of Related Work}
\textbf{Semi-Supervised Object Detection}. 
In SSOD, Pseudo Labeling\cite{self-training} and Consistency-based Regularization\cite{csd},\cite{virtual_consistency} are two commonly used methods. As an early SSOD work, STAC~\cite{stac} proposed a basic multi-stage training framework to combine pseudo labeling and consistency training.
To simplify the multi-stage training process, the end-to-end Teacher-Student framework\cite{mean-teacher},\cite{unbiased-teacher} is proposed, in which the teacher model is updated by exponential moving average (EMA) from the student model, and generates pseudo labels on the unlabeled images in an online manner. Under this framework, a significant amount of research is proposed to improve the quality of pseudo labels~\cite{ismt,instance-teacher,unbiased-teacher,soft-teacher}. Among them, Unbiased Teacher~\cite{unbiased-teacher} replaces the Cross-Entropy loss with Focal Loss to eliminate the class imbalance caused by confirmation bias\cite{bias} of pseudo labels. 
For Consistency-based regularization methods~\cite{pseco, scale_eq}, PseCo~\cite{pseco} introduces the feature-level scale consistency by aligning shifted pyramid features of different scale inputs of the same image.  
Most of these works are based on the two-stage detectors, e.g. Faster RCNN\cite{faster-rcnn}, which involves the anchor generator, a complex hand-crafted component.
On the other hand, some SSOD methods focus on the one-stage detectors~\cite{dsl,dense-teacher,unbiased-v2}. Among them, DSL~\cite{dsl} proposed the first dense learning-based anchor-free SSOD method with adaptive filtering strategy and uncertainty regularization and achieved state-of-the-art performance. Dense Teacher~\cite{dense-teacher} is proposed to use the dense output predictions from the teacher branch as pseudo labels directly to avoid the annoying threshold selection. 
Our Semi-DETR is significantly different from previous works: (1) we explored the challenges of the DETR-based object detectors on SSOD, which, to our best knowledge, is the first systematic research endeavor in SSOD; (2) our Semi-DETR method is tailored for the DETR-based detectors, which eliminates the training efficiency caused by bipartite matching with the noisy pseudo labels and presents a new consistency scheme for set-based detectors.

\section*{Extended Details of Stage-wise Hybrid Matching}
\textbf{Design Details.} In Stage-wise Hybrid Matching, we propose to divide the training process into two stages: the one-to-many assignment training in the first stage and the one-to-one assignment training in the second stage. 

Following the main paper, let us denote the classification score as $s$ and an IoU between the predicted bounding box and the ground truth bounding box as $u$. We take a high-order combination of the classification score and the IoU as the matching score and its negative version as the matching cost in the first stage:
\begin{equation}
  m =  s^{\alpha}\cdot u^{\beta}
  \label{eq:match_criterion_supp}
\end{equation}
\begin{equation}
 \mathcal{C}_{\operatorname{match}}(\hat{y}^t_i, \hat{y}^s_j) =  -m_{ij} = s_{ij}^{\alpha}\cdot u_{ij}^{\beta}
  \label{eq:match_cost_supp}
\end{equation}
where $\hat{y}^t$ and $\hat{y}^s$ are the pseudo labels generated by the teacher and the prediction of the student, respectively.
The $s_{ij}$ is the classification score of $j$-th bounding box prediction to $i$-th ground truth label, and $u_{ij}$ is the IoU between the $j$-th predicted bounding box and the $i$-th ground truth box. The higher the matching score, the better the matching quality between the predicted bounding box and the ground truth box, and the lower the matching cost between them. Then, we assigned multiple positive proposals to each pseudo box according to the matching score as follows:
\begin{equation}
  \hat{\sigma}_{o2m} = 	\left\{\underset{\bm{\sigma_i} \in C^M_{N}}{\arg \min } \sum_{j=1}^{M} \mathcal{C}_{\operatorname{match}}\left(\hat{y}^{t}_{i}, \hat{y}^{s}_{\bm{\sigma_i(j)}}\right)\right\}_{i=1}^{|\hat{y}^t|}.
  \label{eq:one2many_supp}
\end{equation}
where $|\hat{y}^t|$ is the number of the pseudo labels. $C^M_{N}$ is the combination of $M$ and $N$, which denotes that a subset of $M$ proposals is assigned to the pseudo box $\hat{y}^t_i$, and the $\sigma_i(j)$ is the corresponding positive proposal indices. With this assignment strategy, the number of assigned positive proposals for each pseudo label significantly increases, boosting the probability of containing the proposals with higher quality as positive samples and leading to more efficient training. In the implementation, we simply choose the Top-K~(K=M) proposals with the largest matching scores for each pseudo box as the positive proposals.

After the first stage of training, the model is capable to produce high-quality pseudo labels with NMS as the post-process. To enjoy the merit of NMS-free detection without sacrificing the detection performance, we propose to conduct one-to-one assignment training with both the labeled data and unlabeled data in the second stage, where the NMS post-process is applied to the unlabeled data to provide high-quality pseudo boxes. The one-to-one assignment, along with the high-quality pseudo boxes, helps the model to gradually reduce the duplicated predictions and finally evolve into an NMS-free end-to-end detector with better performance.

\begin{figure}[h]
    \centering
    \includegraphics[width=\linewidth]{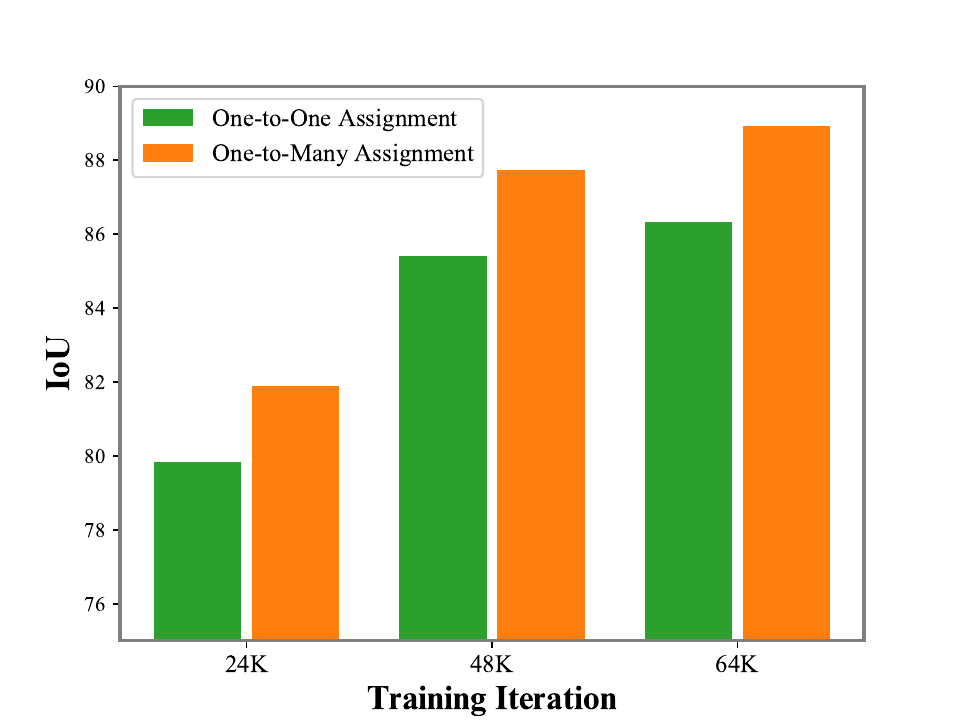}
    \caption{The investigation of the quality of the assigned positive proposals obtained by the one-to-one assignment and one-to-many assignment.}
    \label{fig:comparison_1}
\end{figure}

\textbf{Statistical Analysis.} To validate the effectiveness of our method, we first get the positive candidate proposals obtained by one-to-one assignment using pseudo bounding boxes and the positive candidate proposals obtained by one-to-one assignment using corresponding ground-truth bounding boxes, respectively:
\begin{equation}
\begin{split}
    b^{o2o}_i &= A_{o2o}(b^{pd}_i) \\
\hat{b}^{o2o}_i &= A_{o2o}(b^{gt}_i) \\
\end{split}
  \label{eq:one2one_formulation}
\end{equation}
where $b^{pd}_i$ and $b^{gt}_i$ is the $i$-th pseudo box and its corresponding ground-truth box. $A_{o2o}$ means the one-to-one assignment, i.e. bipartite matching, and the $b^{o2o}_i$ and $\hat{b}^{o2o}_i$ are the corresponding assigned positive proposals. We then calculate the IoU between these two assigned positive proposals:
\begin{equation}
\begin{split}
  I^1_i &= IoU(b^{o2o}_i,\hat{b}^{o2o}_i)
\end{split}
  \label{eq:iou_o2o_formulation_supp}
\end{equation}
The IoU value $I^1_i$ represents the quality of the assigned candidate proposal. The larger the IoU, the closer the assigned positive candidates are to the target object and the better the quality. As a comparison, we get the assigned positive proposals by one-to-many assignment for the $i$-th pseudo box:
\begin{equation}
\begin{split}
  \bm{b^{o2m}_i}=\{b^{o2m}_{i1},b^{o2m}_{i2},...,b^{o2m}_{im}\}= A_{o2m}(b^{pd}_i)
  \end{split}
  \label{eq:one2many_formulation}
\end{equation}
where $A_{o2m}$ is our one-to-many assignment strategy, $\bm{b^{o2m}_i}$ is the multiple assigned positive proposals for $i$-th pseudo box. To verify whether there are positive proposals with higher quality contained in the proposal set obtained by the one-to-many assignment strategy, we calculate the max IoU of these multiple positive proposals and the positive proposal $\hat{b}^{o2o}_i$:
\begin{equation}
\begin{split}
 I^2_i &= Max(\{IoU(b^{o2m}_{i1},\hat{b}^{o2o}_i),...,IoU(b^{o2m}_{im},\hat{b}^{o2o}_i)\})
\end{split}
  \label{eq:iou_o2m_formulation_supp}
\end{equation}
Then, we compare the IoUs $I^1_i$ and $I^2_i$, and the results are shown in Fig.~\ref{fig:comparison_1}. 
It can be found that the multiple positive proposals obtained by our one-to-many assignment strategy clearly contain proposals with higher quality than the proposal obtained by the one-to-one assignment strategy. This result demonstrates that a number of proposals with poor quality are assigned as positive samples due to inaccurate pseudo boxes in the one-to-one assignment, while the correct positive candidate proposals with higher quality are forcibly assigned as negative samples, which finally causes inefficient training.
Our proposed Stage-wise Hybrid Matching applies the one-to-many assignment strategy in the first stage, which enables the proposals with higher quality mistakenly assigned as background to have the opportunity to be optimized. With the modified loss function, the potential positive proposals can be utilized to guide the model convergence while the impact of the proposals with low quality can also be eliminated at the same time.

\begin{figure*}
    \centering
    \includegraphics[width=\linewidth]{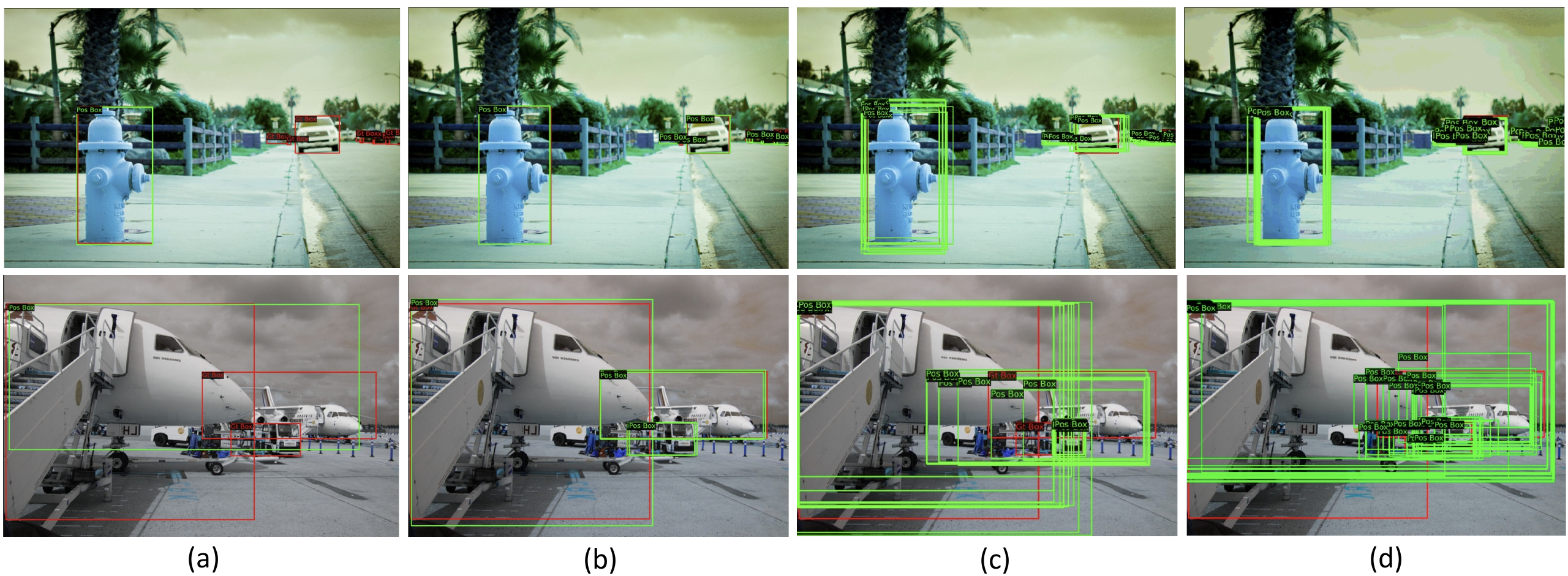}
    \caption{Qualitative results of assigned positive proposals of different one-to-many assignment strategies. (a) Max-IoU(IoU Threshold 0.5) (b) ATSS (c) SimOTA (d) Ours. Compared to Max-IoU, ATSS, and SimOTA, our method obtains more positive proposals for each ground truth bounding box. Note that the ground truth bounding boxes are in red, and the assigned positive predicted bounding boxes are in green.}
    \label{fig:assignment_result_comparison}
\end{figure*}

\begin{figure*}
    \centering
    \includegraphics[width=\linewidth]{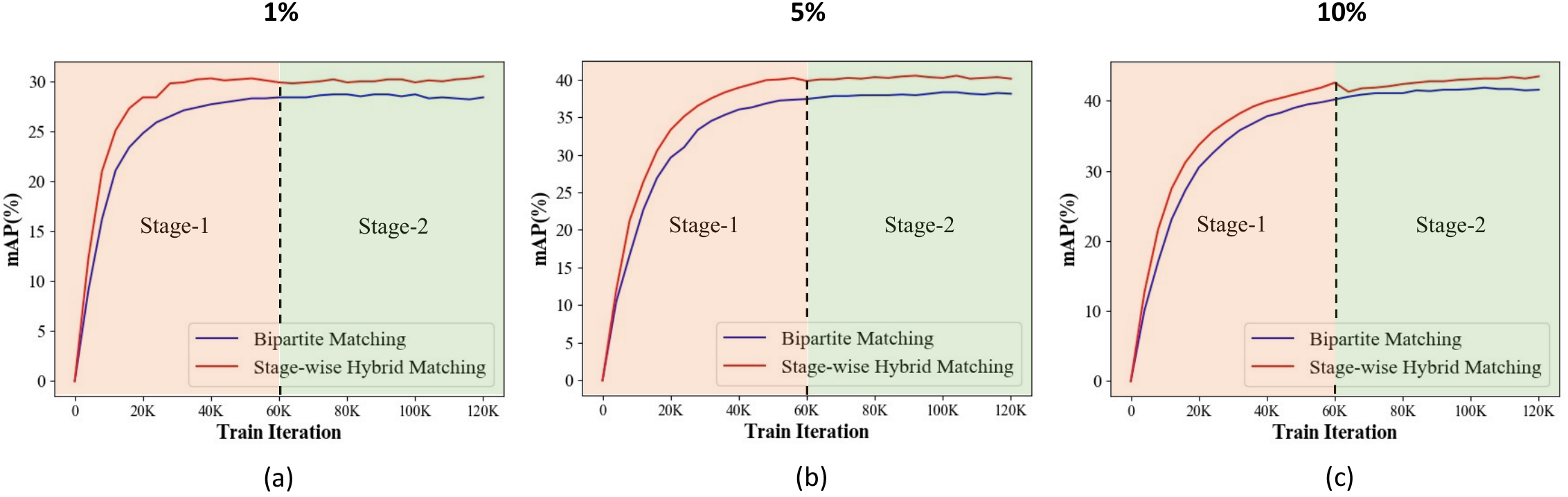}
    \caption{The training efficiency comparisons between the proposed \textbf{State-wise Hybrid Matching}(one-to-many assignment in stage-1 and one-to-one assignment in stage-2) and the original \textbf{Bipartite Matching}(one-to-one assignment) under different labeled data ratios on the COCO dataset. The area in orange and green represent the first stage and second stage in the Stage-wise Hybrid Matching, respectively. Our Stage-wise Hybrid Matching greatly improves the training efficiency, especially when labeled data are scarce.}
    \label{fig:train_curve_comparison}
\end{figure*}

\textbf{Effectiveness Analysis}. As presented in Fig.~\ref{fig:train_curve_comparison}, compared with the one-to-one assignment with Bipartite Matching\cite{detr}, the proposed Stage-wise Hybrid Matching greatly improves the training efficiency of the first stage thanks to the multiple assigned positive proposals.
More importantly, the performance improvement becomes more prominent when the number of labeled data gets more scarce, e.g., 1\%,  which demonstrates the superiority of our method. Furthermore, we compare different alternative one-to-many assignment strategies in our method. Specifically, we replace the one-to-many assignment strategy used in our method with the Max-IoU\cite{faster-rcnn}, ATSS\cite{atss}, and SimOTA\cite{yolox}, respectively. We conduct these experiments with Semi-DETR with DINO\cite{dino}, and all the models are trained for 60K iterations. The results are shown in Tab.~\ref{ablation_assignment_method}. Interestingly, although designed to assign multiple positive proposals, both Max-IoU and ATSS do not perform well in DETR-based detectors. We visualize the assignment results of these assignment strategies in Fig.~\ref{fig:assignment_result_comparison}. 
When applying the Max-IoU assignment strategy, we observe that only a few ground truth boxes own lots of duplicated positive proposals, and most ground truth boxes have no positive assigned proposals.  We suspect the main reason is that the learnable object queries are unable to always guarantee enough IoU with each ground truth box, and constantly changed during the training.  Unlike the fixed anchor box prior, the predicted proposals of the object queries easily cluster around a particular ground truth box, and finally leads to duplicated positive proposals, which is not helpful for the training. However, further discussion is beyond this paper. As a comparison, the ATSS assignment strategy generates a few positive proposals for each ground truth box. However, the number of positive proposals obtained by the ATSS for each ground truth box is still limited. This is because the ATSS only considers the IoU during the assignment, and the adaptive IoU thresholds obtained by the ATSS are so high that most of the possible high-quality proposals are filtered out.
Different from the Max-IoU and ATSS, both SimOTA\cite{yolox} and our proposed method achieve much better performance, which benefits from (1) the ranking-based one-to-many assignment strategy via the top-K operation to ensure enough positive proposals for each ground truth box, and (2) the ranking criteria considering both the classification score and IoU score, which can generate various positive proposals as shown in \ref{fig:assignment_result_comparison}.

\begin{table}[h]
\centering
\caption{Performance comparisons among different one-to-many assignment strategies. Baseline means the one-to-one assignment with bipartite matching. All the models are trained for 60K iterations.}
\begin{tabular}{c|c|c|c}
\hline  
Method& mAP& $AP_{50}$ & $AP_{75}$\\
\hline  
Baseline&40.2&56.5&43.4\\
Max-IoU&11.4&15.0&12.1\\
ATSS & 18.7&30.5&18.9\\
SimOTA& 42.5&\textbf{59.9}&45.2\\
Ours& \textbf{42.8}&59.8&\textbf{46.0}\\
\hline 
\end{tabular}
\label{ablation_assignment_method}
\end{table}

\section*{Extended Details of Cross-view Query Consistency}
\begin{figure*}
    \centering
    \includegraphics[width=0.3\linewidth]{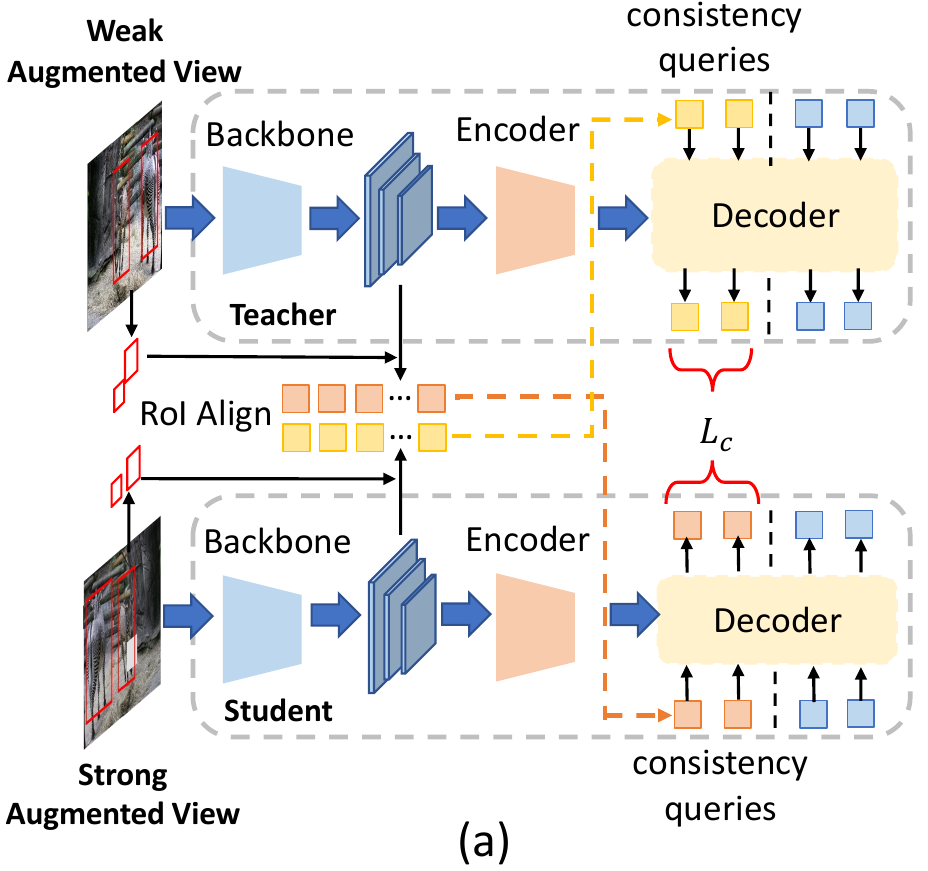}
    \includegraphics[width=0.3\linewidth]{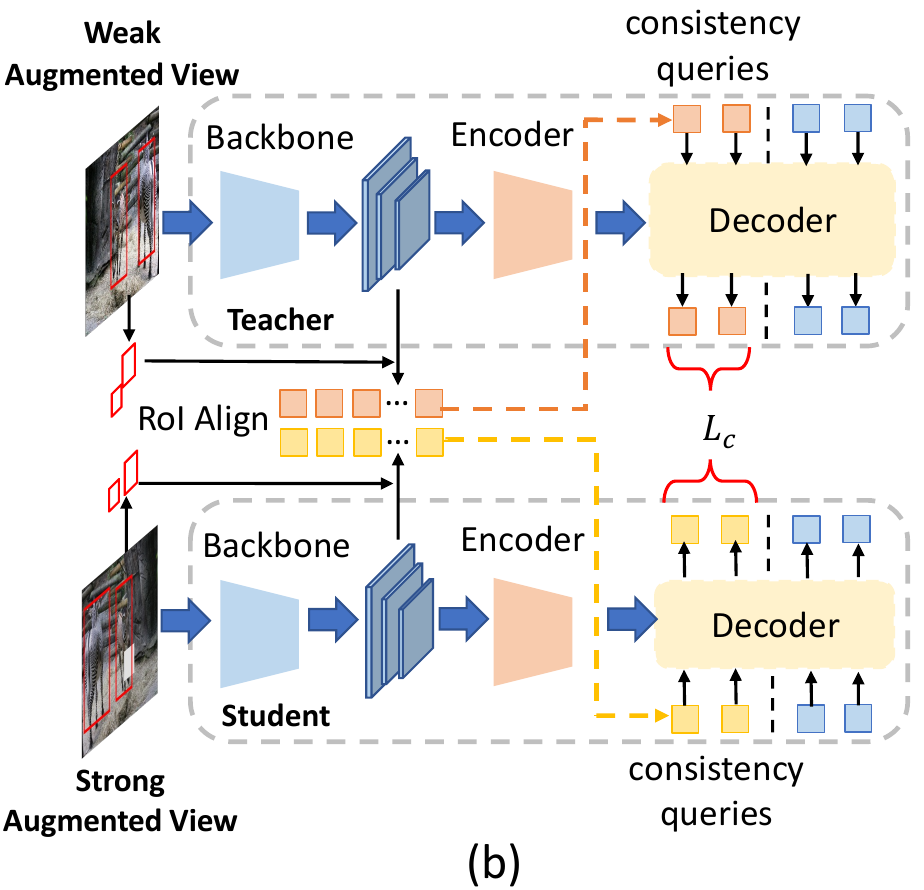}
    \includegraphics[width=0.3\linewidth]{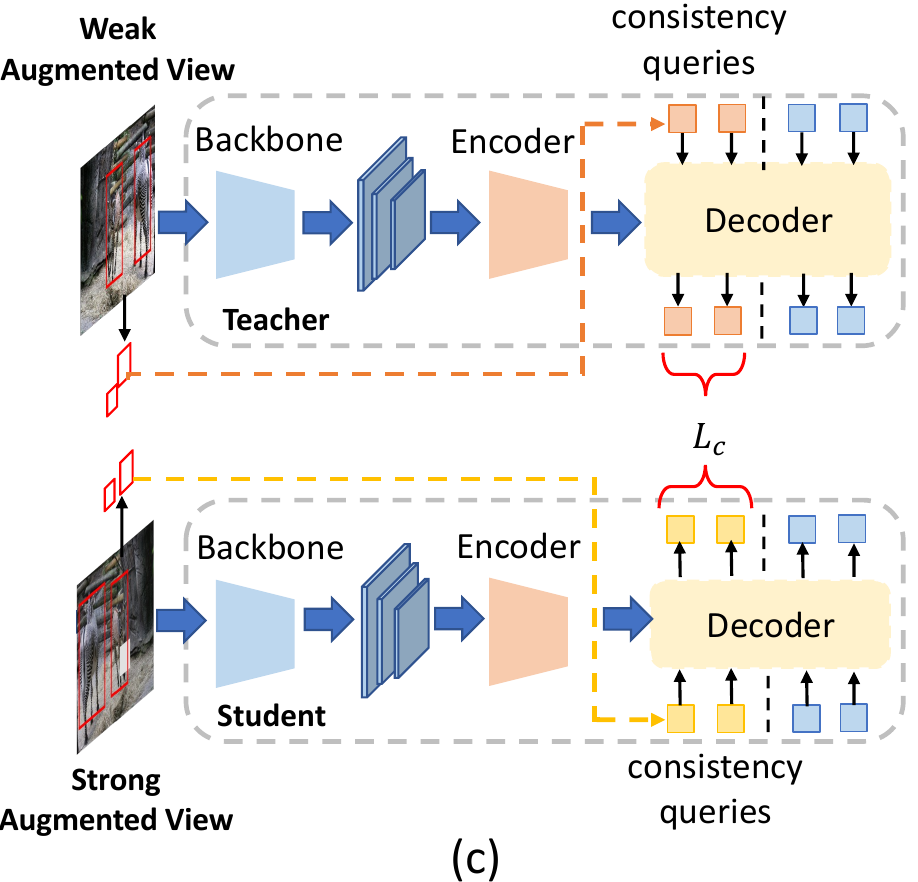}
    \caption{Overview of different variants of the cross-view query consistency. (a) our proposed cross-view query consistency, (b) consistency scheme in Exp-1 which replaces the cross-view consistency queries with single-view consistency queries, and (c) consistency scheme in Exp-2, which directly takes the positional embedding of pseudo boxes as the consistency queries.}
    \label{fig:consistency_variant}
\end{figure*}

We conduct experiments to validate the effectiveness of our cross-view query consistency. We aim to answer the following two questions:
\begin{enumerate}[noitemsep, nolistsep]
  \item Whether the cross-view queries necessary? Can we replace the cross-view queries with single-view queries?
  \item Whether the RoI features in the cross-view queries really matter? What about conducting consistency training without incorporating these features into the consistency queries?
\end{enumerate}
To answer these two questions, we conduct two experiments as follows:
\textbf{Exp-1}: We construct the consistency queries by the RoI features within each view separately. And then, we perform the query decoding in the teacher and student models individually. Finally, we impose the consistency constraint on the consistency queries decoding embedding of the teacher and student models. The overview is presented in Fig.~\ref{fig:consistency_variant}(b). As shown in Tab.~\ref{ablation_consistnecy_method}, compared with our proposed cross-view query consistency, when replacing the cross-view queries with single-view queries, the performance shows a 0.5 mAP drop. It confirms the importance of the cross-view queries in our consistency scheme. The possible reason for the effectiveness of these cross-view queries is that these queries provide information about the object from another view, which encourages learning the semantic invariance during decoding and leads to better performance.

\textbf{Exp-2}: As shown in Fig.~\ref{fig:consistency_variant}(c), we construct the consistency queries directly based on the positional embedding of the pseudo boxes without the RoI features of corresponding pseudo boxes. The difference between this scheme and DN-DETR\cite{dn-detr} is that we do not add the noise into the pseudo boxes before obtaining the positional embedding. The results are presented in Tab.~\ref{ablation_consistnecy_method}. After removing the RoI features during the construction of the consistency queries, the performance greatly decreased to 42.7(-0.8) mAP, which demonstrates the necessity of the RoI features in consistency queries. The reason for this performance degeneration is that the positional embedding of the pseudo boxes does not have strong priors to guarantee the correspondence between the consistency queries input and their corresponding output prediction, which increases the learning difficulty of the consistency training. As a comparison, we take the RoI features from different views as the strong semantic guidance during the decoding and ensure the final decoder embedding is relevant to the input consistency queries, which eventually leads to the success of the consistency training.

\begin{table}[h]
\centering
\caption{Performance comparisons of different variants of Cross-view Query Consistency}
\begin{tabular}{c|c|c|c}
\hline  
Method& mAP& $AP_{50}$ & $AP_{75}$\\
\hline  
Exp-1&43.0&59.3& 46.3\\
Exp-2&42.7&58.9&46.0\\
Ours & \textbf{43.5}&\textbf{59.7}&\textbf{46.8}\\
\hline 
\end{tabular}
\label{ablation_consistnecy_method}
\end{table}

\section*{Extended Details of the Cost-based Pseudo Label Mining}

\textbf{Design Details.} Concretely, we take two steps to generate the pseudo boxes for consistency training with a good trade-off between precision and recall. First, for each unlabeled image, we calculate the mean $\mu$ and variance $\sigma$ of the confidence scores of the detection results. Then, we take the threshold $\tau_1 = \mu + \sigma$ to filter and get the initial pseudo boxes. 

For the second step, we perform the bipartite matching with these initial pseudo boxes and the student model predicted proposal boxes, and record the matching cost of each pseudo box.

\begin{figure*}[ht]
	\centering
	\begin{minipage}[t]{\linewidth}
		\centering
		\includegraphics[width=8cm]{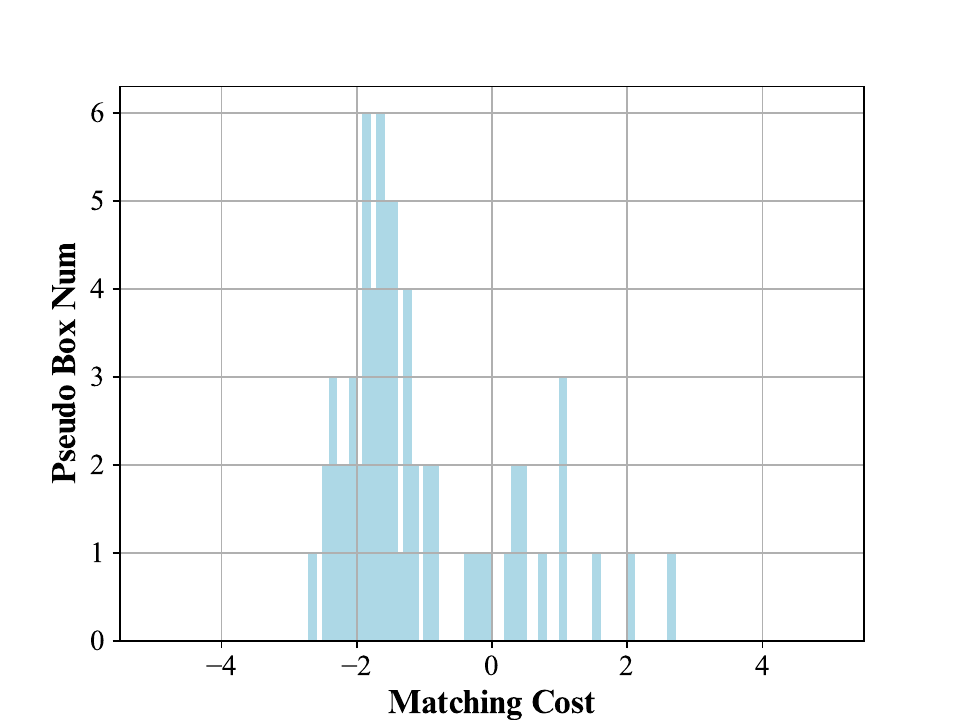} 
		\includegraphics[width=8cm]{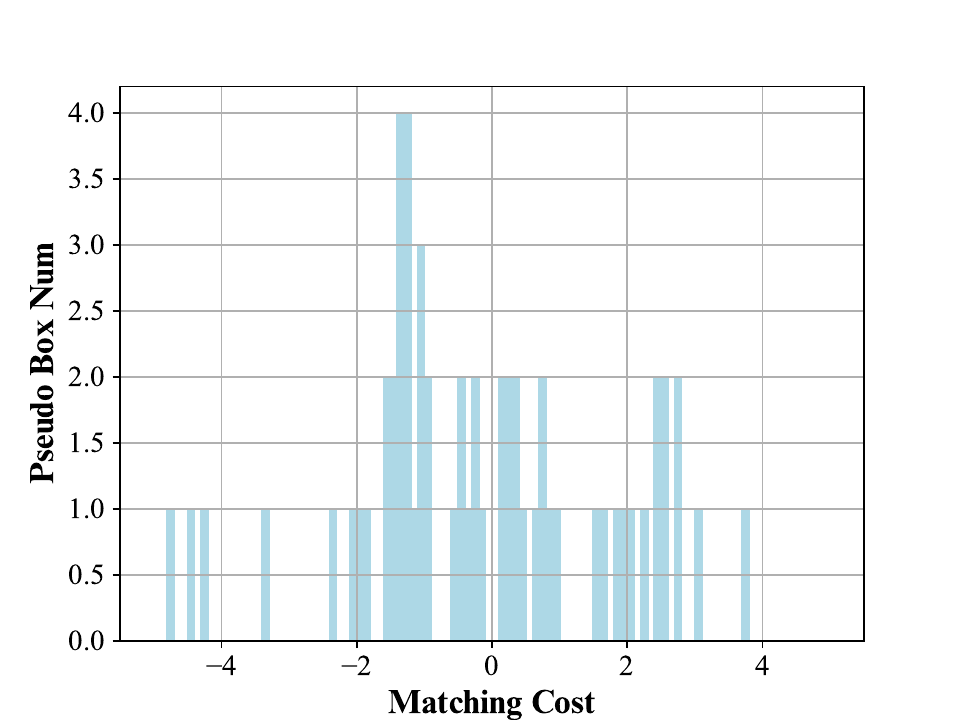} 
	\end{minipage}
	\begin{minipage}[t]{\linewidth}
		\centering
		\includegraphics[width=8cm]{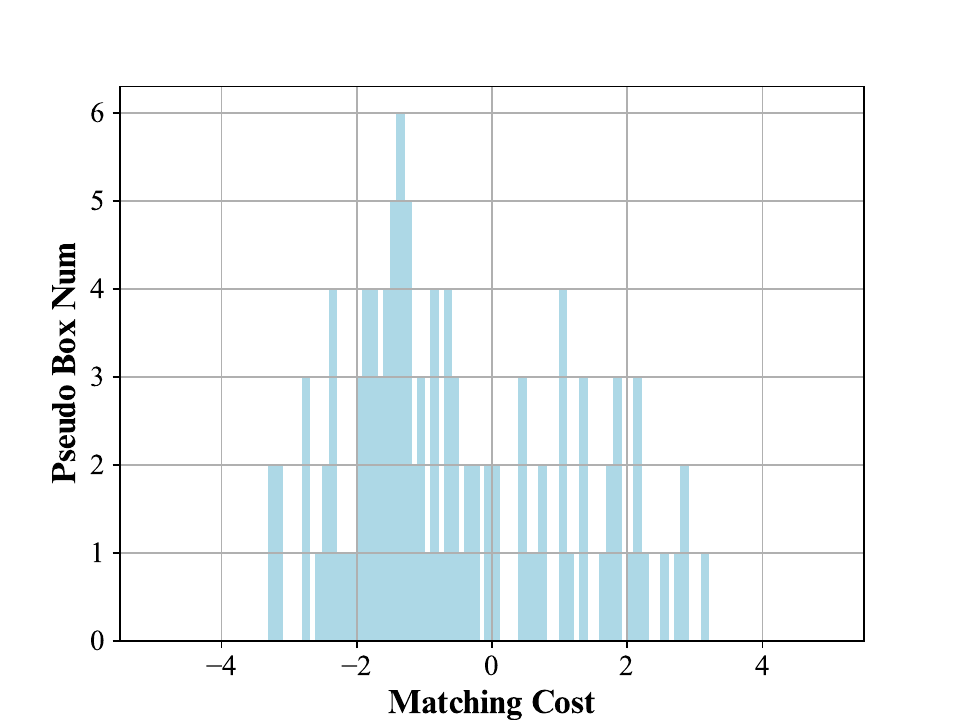} 
		\includegraphics[width=8cm]{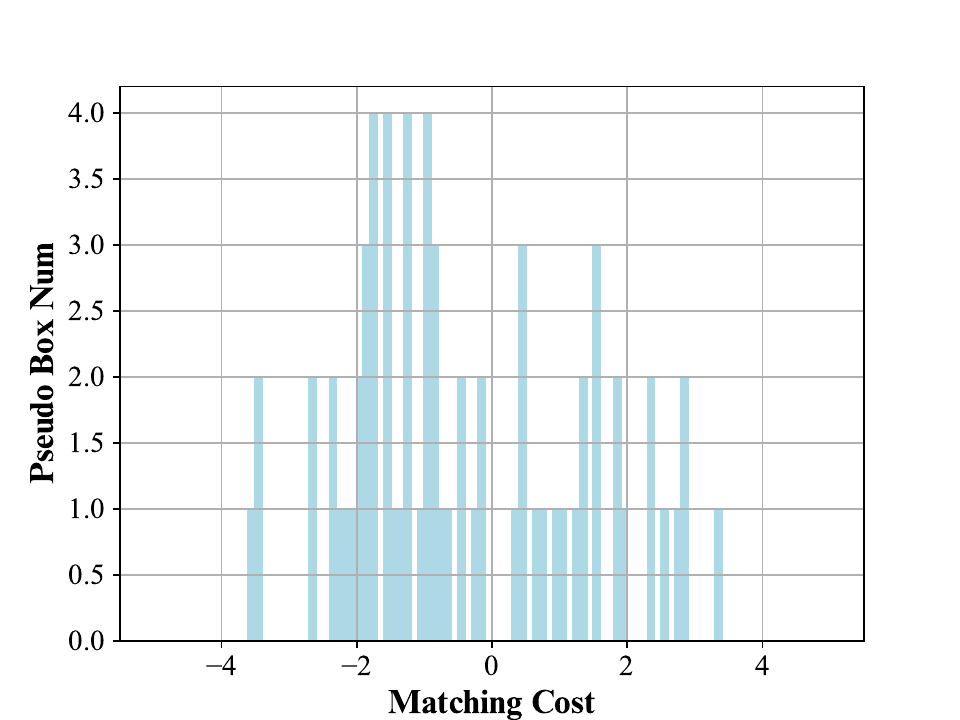} 
	\end{minipage}
	\caption{The distribution of the matching cost of the initial pseudo boxes within a random batch.}
	\label{fig_matching_cost_distritbution}
\end{figure*}

We collect the matching cost of the initial pseudo boxes within a batch and show the distribution of the matching cost in Fig.~\ref{fig_matching_cost_distritbution}. Obviously, the distribution of the matching costs presents a bimodal distribution.

\begin{figure*}[htp]
	\centering
	\begin{minipage}[t]{\linewidth}
		\centering
		\includegraphics[width=7cm, height=5cm]{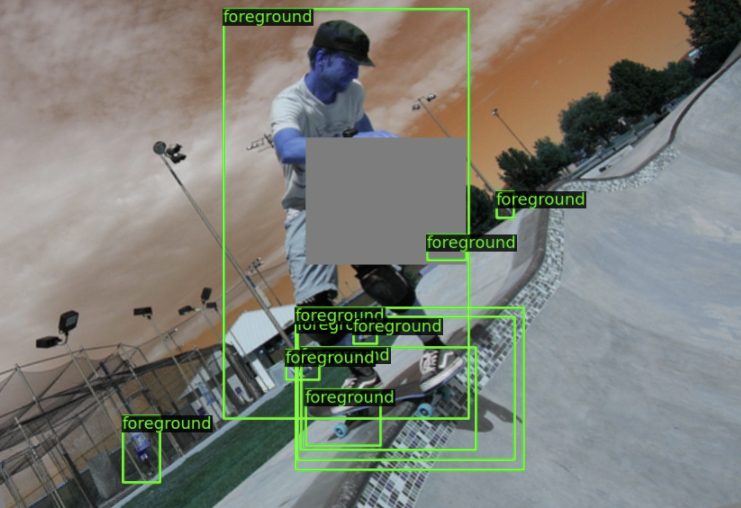} 
		\includegraphics[width=7cm, height=5cm]{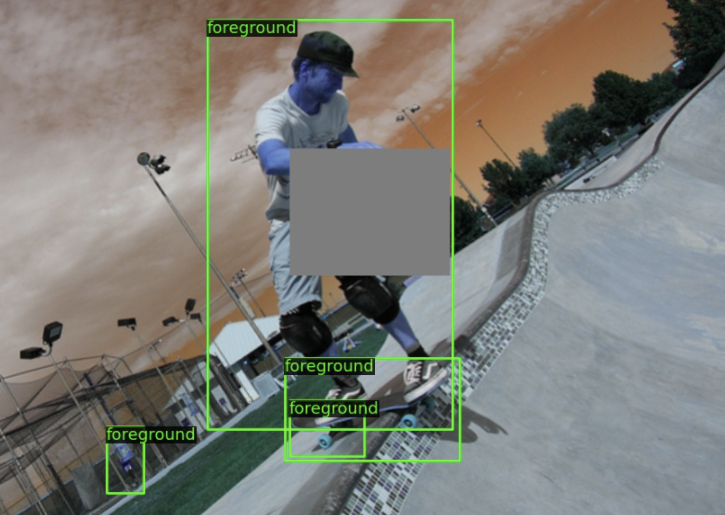} 
	\end{minipage}
	\begin{minipage}[t]{\linewidth}
		\centering
		\includegraphics[width=7cm, height=5cm]{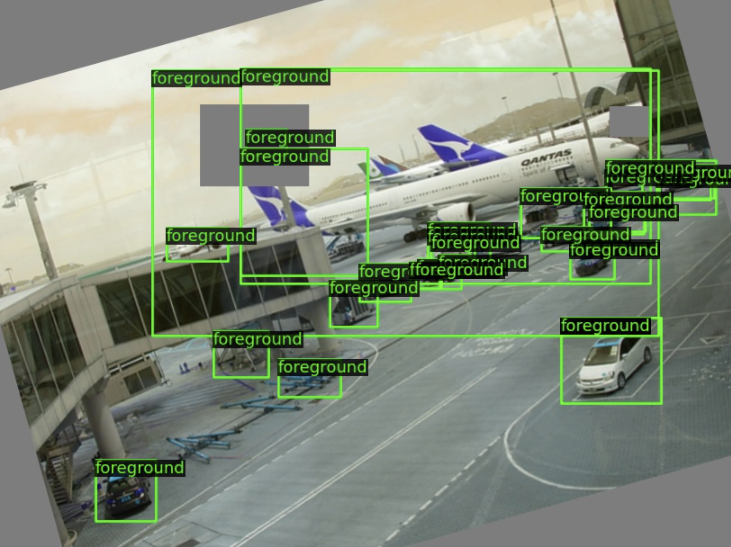} 
		\includegraphics[width=7cm, height=5cm]{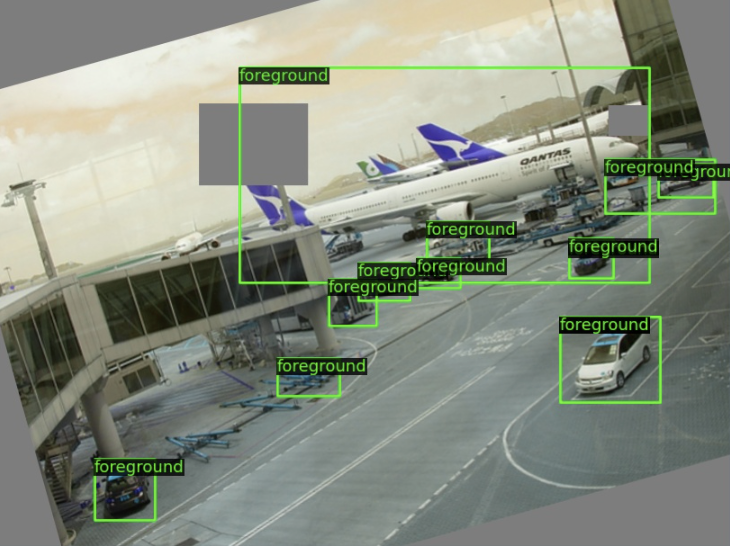} 
	\end{minipage}
	\begin{minipage}[t]{\linewidth}
		\centering
		\includegraphics[width=7cm, height=5cm]{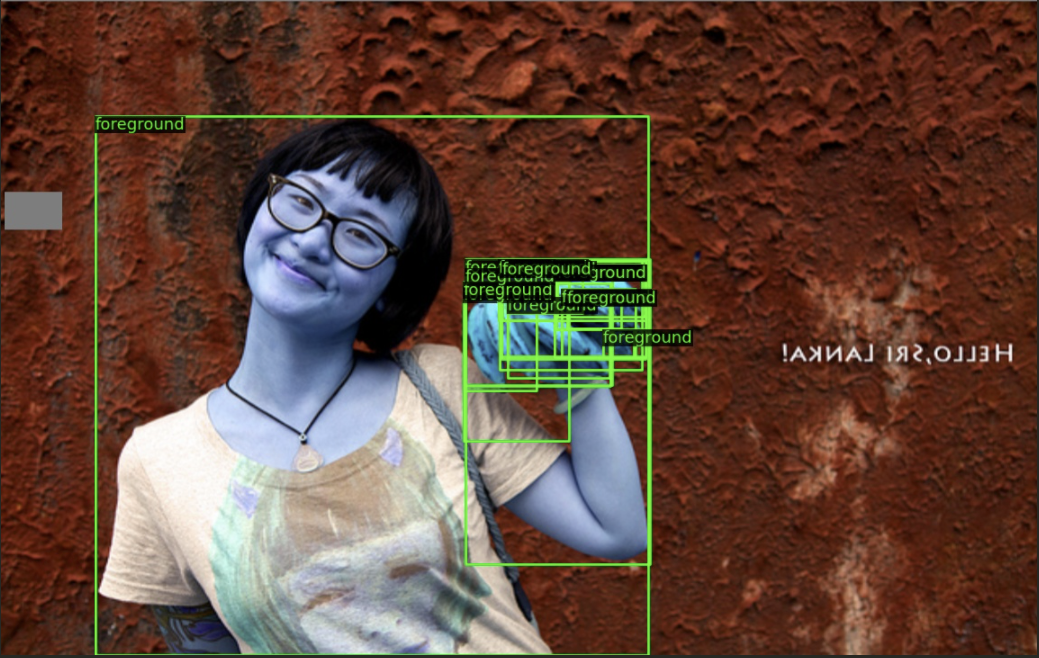} 
		\includegraphics[width=7cm, height=5cm]{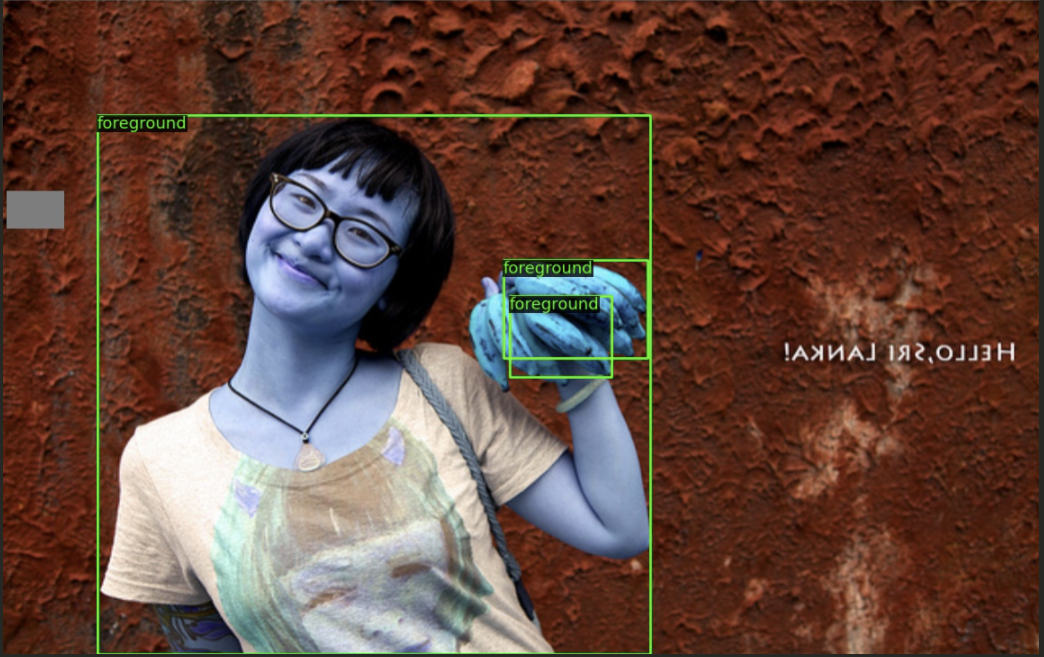} 
	\end{minipage}
	\begin{minipage}[t]{\linewidth}
	\centering
	\includegraphics[width=7cm, height=5cm]{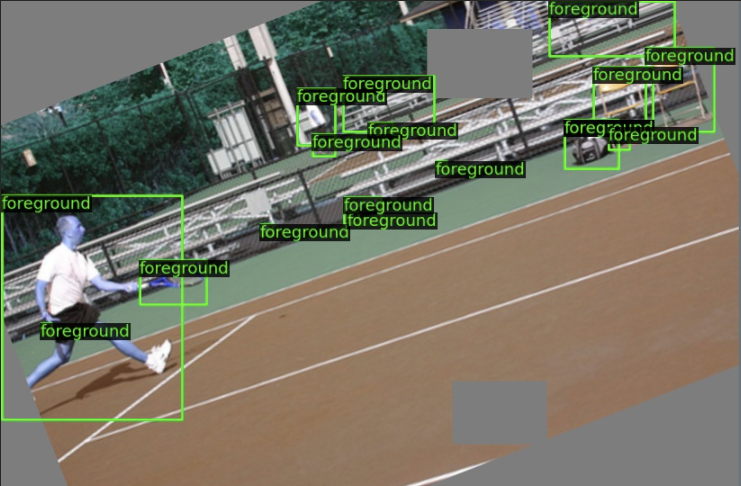} 
	\includegraphics[width=7cm, height=5cm]{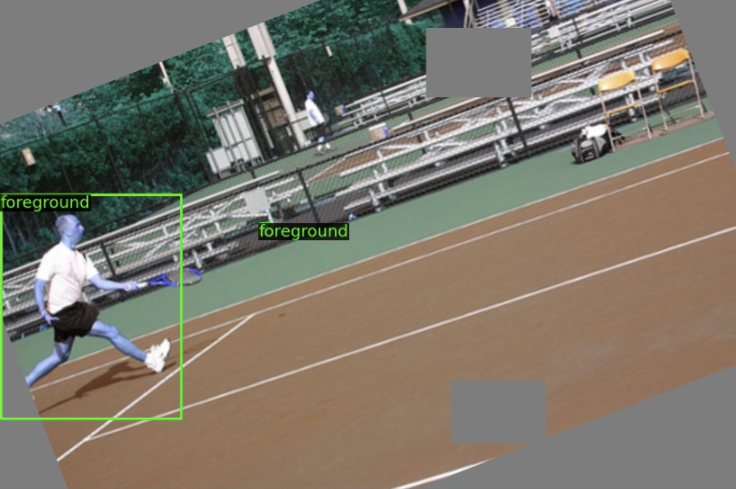} 
	\end{minipage}
	\caption{The pseudo boxes before and after the cost-based pseudo label mining. \textbf{Left}: the initial pseudo boxes obtained with threshold $\tau_1$, \textbf{Right}: the pseudo boxes after the cost-based pseudo label mining with GMM. Note that we applied strong augmentations on the unlabeled images and visualized the predictions accordingly.}
	\label{result_gmm_filtering}
\end{figure*}

To this end, we propose to model the cost distribution with a Gaussian Mixture Model(GMM) consisting of two Gaussian distributions as follows:
\begin{equation}
    P(c|\theta) = w_r\mathcal{N}_{r}(c,\mu_r,\sigma_r) + w_u\mathcal{N}_u(c,\mu_u,\sigma_u)
    \label{eq:gmm_model}
\end{equation}
where the $P(c|\theta)$ means the probability of matching cost value $c$, $\theta$ is the parameters of the GMM model. $\mathcal{N}_{r}(c,\mu_r,p_r)$ represents the cost distribution of reliable pseudo boxes with lower matching cost and $\mathcal{N}_u(c,\mu_u,p_u)$ represents the cost distribution of unreliable pseudo boxes with higher matching cost. $w_r$ and $w_u$ represent the blending weights of these two distributions, $\mu_r$(or $\mu_u$), and $\sigma_r$(or $\sigma_u$) represent the means and variances of these two distributions, respectively. 
The fitting process can be solved by the Expectation-Maximization~(EM) algorithm~\cite{em}. Then, we set the threshold $\tau_c$ as the cost with the highest probability of being the reliable pseudo boxes.
\begin{equation}
\tau_c = \underset{c}{\arg \max }P_{reliable}(c|c,\theta)
  \label{eq:gmm_thr}
\end{equation}
The bounding boxes with matching costs less than $\tau_c$ are regarded as reliable pseudo boxes and are retained for cross-view query consistency learning. As shown in Fig.~\ref{result_gmm_filtering}, this pseudo label mining method successfully mines more pseudo bounding boxes which is reliable for consistency training from the initial pseudo boxes.

\textbf{Effectiveness Analysis.} In our main paper, we take the fixed classification confidence score $\tau_s = 0.4$ to filter and obtain the pseudo labels for the training of classification and regression. The pseudo labels obtained by cost-based pseudo label mining (CPM) are used for consistency training only. Here, we conduct experiments to test the extension of the CPM to replace this fixed threshold filtering scheme. 

According to Tab.~\ref{ablation_pseudo_label_generations}, interestingly, when the pseudo labels from the CPM are utilized to train the classification and regression losses, the detector suffers a clear performance drop (-1.1\%). 

This indicates that these pseudo labels are not suitable for the training of classification and regression. The possible reasons for this performance drop are two-fold: (1) our proposed Cross-view Query Consistency aims to learn semantic feature invariance between different views from unlabeled images, which essentially does not have a strict requirement for the localization accuracy(i.e. high precision) of the pseudo bounding boxes. Meanwhile, the CPM generates more pseudo bounding boxes(i.e. high recall) than that of the fixed threshold filtering scheme, which essentially facilitates the learning of cross-view query consistency. (2) As discussed in the Stage-wise Hybrid Matching part in the main paper, the one-to-one assignment strategy used in DETR-based detectors requires more accurate(i.e. high precision) pseudo labels to effectively supervise the classification and regression learning, otherwise would lead to inefficient training.

\section*{Data Augmentations}
Generally, we follow the data augmentation scheme in Soft-Teacher\cite{soft-teacher}. We summarize the data augmentations used in our method in Tab.~\ref{table_augmentation}. Note that we do not use more advanced data augmentations such as Large Scale Jittering in \cite{soft-teacher,dense-teacher}, MixUp \cite{mixup}, and Mosaic in \cite{instance-teacher}, Patch Shuffle in \cite{dsl}. We believe these data augmentations can further improve our performance, which we leave for future work.
\begin{table}[h]
\centering
\caption{Experiments about usage extension of the pseudo labels from Cost-baed Pseudo Label mining(CPM). Cls means classification training and Reg means regression training. Consistency represents the cross-view query consistency.}
\resizebox{\linewidth}{!}{
\begin{tabular}{c|c|c|c}
\hline  
Method&  Cls + Reg & Consistency& mAP\\
\hline  
CPM(Ours)&&\checkmark& \textbf{43.5}\\
CPM(Extension)&\checkmark&\checkmark&42.4\\
\hline 
\end{tabular}}
\label{ablation_pseudo_label_generations}
\end{table}

\section*{Extended Details of Experiments}
Here, we provide more details about the experiments with Deformable DETR\cite{df-detr}, i.e. Semi-DETR(Def-DETR). (1) For the COCO Partial benchmark, we train Semi-DETR(Def-DETR) for 180k iterations and the training time of the first stage with one-to-many assignment $T_1$ is set to 120k iterations. Other settings are kept the same with Semi-DETR(DINO). (2) For the COCO-Full benchmark, the total training time is set to 240k iterations, and $T_1$ is set to 180k iterations. Other settings are kept the same with Semi-DETR(DINO). (3) For the Pascal VOC benchmark, we train Semi-DETR(Def-DETR) for 120k iterations with the training time of first stage $T_1$ set to 80k iterations. Other settings are kept the same with COCO-Partial benchmark. For all experiments, the confidence threshold is set to 0.4. We utilize Adam\cite{adam} with a learning rate of 2e-4 and weight decay of 0.0001, and no learning rate decay scheme is used. The teacher model is updated from the student model through EMA with a momentum of 0.999. 

\textbf{Comparisons to Omni-DETR}. Omni-DETR~\cite{omni} is a DETR-based object detector designed for omni-supervised object detection. It is not designed specifically for SSOD as admitted in their paper, but it is extended to the SSOD task by introducing a simple pseudo-label filtering scheme. Our Semi-DETR is significantly different from Omni-DETR in the following aspects: 

(1) \textbf{Different motivations for model design}. To perform SSOD, Omni-DETR adopted simple hard thresholding on the confidence scores of the predictions to assign supervised pseudo-labels to unlabeled data, which can be viewed as a simple integration of DETR-based detectors to the general SSOD framework. We conducted an in-depth analysis of this pipeline and identified that the one-to-one assignment strategy leads to training inefficiency due to inaccurate pseudo labels, especially during the early training phase. Besides, the lack of deterministic correspondence between the input query and its prediction output in DETR-based detection framework also hinders the integration of consistency-based regularization which is known to be effective in existing SSOD methods. Consistency-based regularization is therefore not explored in Omni-DETR. Our proposed Semi-DETR alleviates the training inefficiency by combining the one-to-many and one-to-one assignment strategies to provide pseudo-labels of higher quality. Moreover, it introduces a consistency-based regularization scheme powered by a cost-based pseudo label mining method, which enables consistency regularization for DETR-based detectors. In general, compared to Omni-DETR, Semi-DETR is a tailored design for SSOD, and it is an important step forward to extend the study of DETR-based detectors to SSOD.

(2) \textbf{Different training strategy}. Omni-DETR follows the complex multi-stage training pipeline of Unbiased-Teacher\cite{unbiased-teacher}, which requires an extra burn-in stage to pre-train on labeled data and thus \textbf{is not essentially end-to-end}. However, Semi-DETR shares the same design philosophy as Soft-Teacher\cite{soft-teacher} without the need to pre-train with labeled data in advance. Both detectors embrace the benefits of NMS-free post-process, but our proposed Semi-DETR achieves end-to-end detection in both the training and inference phases. \textbf{This strengthens our claim that Semi-DETR is the first transformer-based end-to-end semi-supervised object detector}.

(3) \textbf{Significant performance improvement}. As discussed in~\cite{omni}, Omni-DETR utilizes Deformable DETR as the base detector for faster convergence. We compare our Semi-DETR with Omini-DETR using the same baseline detectors under different COCO-Partial settings as in Tab.~\ref{ablation_detector}. Clearly, Semi-DETR achieves SOTA performance with different detectors, and it is superior to Omni-DETR across all base detectors under different experimental settings. 

We present fair comparisons between Omni-DETR and Semi-DETR using different base detectors~(i.e., Deformable DETR and DINO) in Tab.~\ref{ablation_detector}. \textbf{First we must clarify that Omni-DETR actually adopts Deformable DETR as the base detector due to the slow convergence of original DETR}. The performance of Omni-DETR with Deformable DETR is thus directly copied from \cite{omni}, and we additionally evaluate its performance with DINO. As shown in Tab.~\ref{ablation_detector}, Semi-DETR consistently achieves better performance than Omni-DETR across all settings. Moreover, \textbf{even armed with DINO as the base detector, our Semi-DETR still outperforms Omni-DETR by clear margins}, which manifests the superiority of our Semi-DETR in terms of performance compared to Omni-DETR. 

\begin{table}[h]
\centering
\caption{Performance comparisons between Omni-DETR and Semi-DETR with different detectors under COCO-Partial settings. 
}
\resizebox{\linewidth}{!}{
\begin{tabular}{c|c|c|c}
\hline  
Method& 1\% & 5\% & 10\%\\
\hline  
Omni-DETR(Def-DETR)&18.60&30.20& 34.10\\
Semi-DETR(Def-DETR)&\textbf{25.20} &\textbf{34.50}&\textbf{38.10}\\
 \textit{Improvement}& {\color{blue} +6.60}&{\color{blue} +4.30}&{\color{blue} +4.00} \\
 \hline
Omni-DETR(DINO)& 27.60&37.70&41.30\\
Semi-DETR(DINO)& \textbf{30.50}&\textbf{40.10}&\textbf{43.50}\\
 \textit{Improvement}& {\color{blue} +2.90}&{\color{blue} +2.40}&{\color{blue} +2.20} \\
\hline 
\end{tabular}}
\label{ablation_detector}
\end{table}

\section*{More Visualization}
\begin{figure*}[h]
    \centering
    \includegraphics[width=0.72\linewidth]{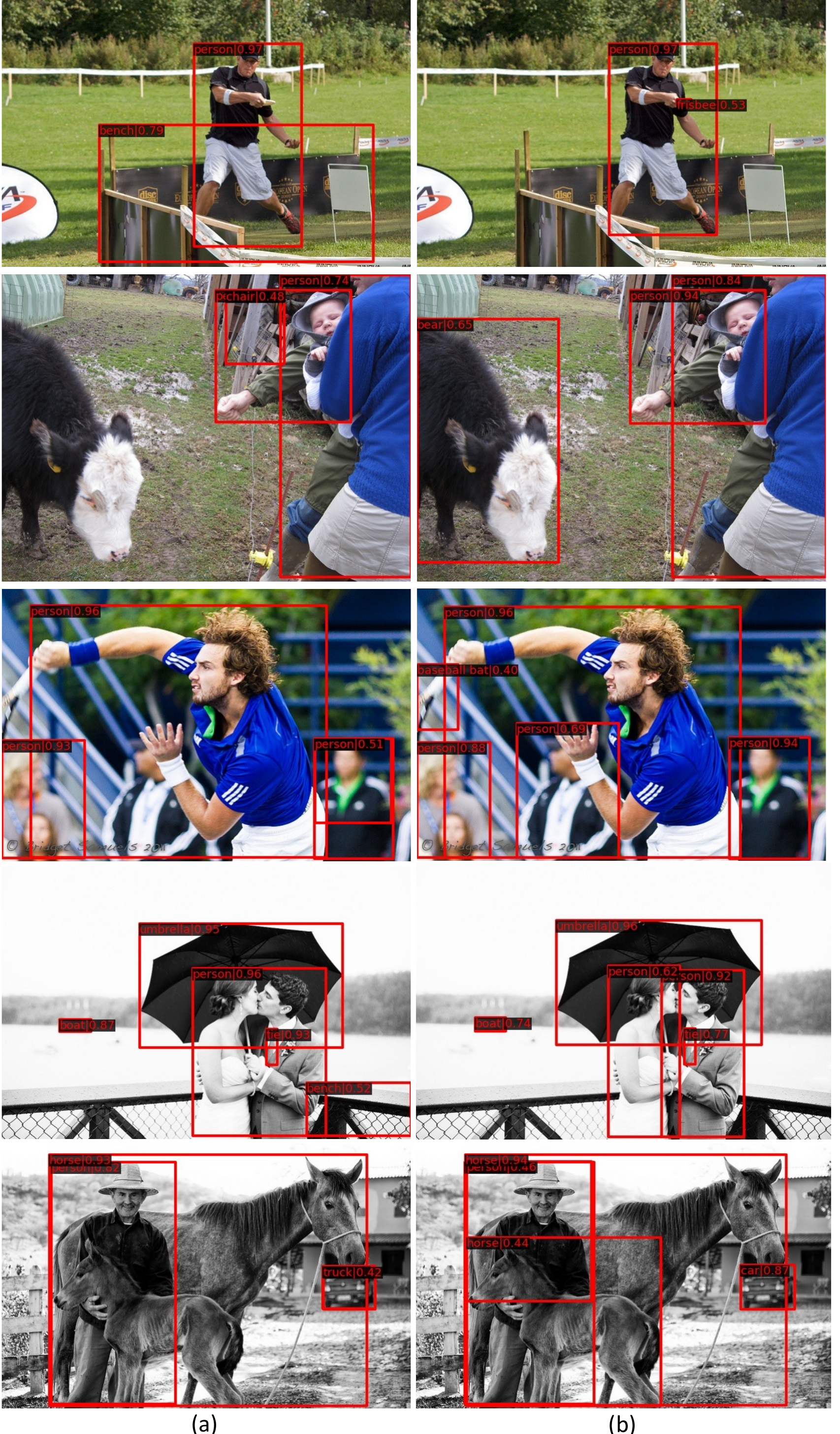}
    \caption{Qualitative comparisons between the pseudo labels generated by (a) training without Stage-wise Hybrid Matching and (b) training with Stage-wise Hybrid Matching. The pseudo label threshold  $\tau_s=0.4$.}
    \label{fig:pseudo_label_visualization}
\end{figure*}

Stage-wise Hybrid Matching improves the training efficiency when the pseudo labels are inaccurate by the one-to-many assignment, and makes it able to generate the pseudo labels with higher quality in the second stage. To validate this, we visualize the pseudo labels training with and without the State-wise Hybrid Matching. The results are shown in Fig.~\ref{fig:pseudo_label_visualization}. It clearly shows that our Stage-wise Hybrid Matching generates better pseudo boxes.

\section*{Discussion of Limitations}
Achieving an end-to-end detection framework without NMS post-processing under DETR-based semi-supervised object detection, while maintaining the performance of a fully one-to-many assignment strategy is a research direction worth exploring. Semi-DETR has demonstrated the effectiveness of combining the one-to-many assignment and the one-to-one assignment strategies at the cost of a performance drop compared to the fully one-to-many assignment strategy. Nevertheless, how to design a better DETR-based SSOD framework that could minimize this performance gap remains an open problem in the research community. We leave it to future work.


\begin{table*}[ht]
\centering
\caption{Data augmentations used in our method. p represents the probability of choosing a certain type of augmentation.}
\begin{tabular}{c|c|c|c}
\hline  
Augmentation & Labeled image training & Unlabeled image training & Pseudo-label generation\\ 
\hline 
 Scale Jitter &
 shortest edge $\in$ $[480, 800]$
  & shortest edge $\in$ $[480, 800]$ & shortest edge $\in$ $[480, 800]$ \\

Solarize Jitter&p=0.25,ratio$\in$(0,1)&p=0.25,ratio$\in$(0,1) & -\\

Brightness&p=0.25,ratio$\in$(0,1)&p=0.25,ratio$\in$(0,1) & -\\

Constrast Jitter& p=0.25, ratio $\in$(0,1)& p=0.25, ratio $\in$(0,1)& -\\
Sharpness Jitter& p=0.25, ratio $\in$(0,1)& p=0.25, ratio $\in$(0,1) & -\\
Translation& -&p=0.3, translation ratio$\in$(0,1)& -\\
Rotate & - &p=0.3,angle$\in$(0,\ang{30})&-\\
Shift&-&p=0.3,angle$\in$(0,\ang{30})&-\\
Cutout&num$\in$(1,5),ratio$\in$(0.05,0.2)&num$\in$(1,5),ratio$\in$(0.05,0.2)&- \\
\hline 
\end{tabular}
\label{table_augmentation}
\end{table*}


\end{document}